\documentclass[sn-mathphys-num]{sn-jnl}

\usepackage{graphicx}%
\usepackage{multirow}%
\usepackage{amsmath,amssymb,amsfonts}%
\usepackage{amsthm}%
\usepackage{ulem}
\usepackage{amsmath}
\usepackage{mathrsfs}%
\usepackage[title]{appendix}%
\usepackage{xcolor}%
\usepackage{textcomp}%
\usepackage{manyfoot}%
\usepackage{booktabs}%
\usepackage{algorithm}%
\usepackage{algorithmicx}%
\usepackage{algpseudocode}%
\usepackage{listings}%
\usepackage{array}%
\usepackage{algorithm}
\usepackage{algpseudocode}
\usepackage{pifont}%
\usepackage{utfsym}
\usepackage{bbding}
\usepackage{xcolor} %
\usepackage{ulem} %

\usepackage{geometry}
\usepackage{tabularx}
\usepackage{booktabs}
\usepackage{multirow}
\usepackage{xcolor}
\usepackage{multicol}
\usepackage{listings}
\theoremstyle{thmstyleone}%
\theoremstyle{thmstyletwo}%

\theoremstyle{thmstylethree}%

\begin{document}

\title[Article Title]{Chat with UAV -- Human-UAV Interaction Based on Large Language Models}


\author[1,4]{\fnm{Haoran} \sur{Wang}}

\author[1,2]{\fnm{Zhuohang} \sur{Chen}}

\author[1,2]{\fnm{Guang} \sur{Li}}

\author*[2]{\fnm{Bo} \sur{Ma}}\email{mabo@mail.zjgsu.edu.cn}

\author[3]{\fnm{Chuanhuang} \sur{Li}}



\affil[1]{\orgdiv{School of Engineering and Informatics}, \orgname{University of Sussex}, \orgaddress{\street{Falmer}, \city{Brighton}, \postcode{BN1 9RH}, \country{England}}}

\affil[2*,3]{\orgdiv{School of Information and Electronic Engineering}, \orgname{Zhejiang Gongshang University}, \city{Hangzhou}, \postcode{310018}, \state{Zhejiang}, \country{China}}

\affil[4]{\orgdiv{Department of Electrical and Computer Engineering}, \orgname{University of Auckland}, \city{Auckland}, \postcode{1142}, \country{New Zealand}}



\abstract{ The future of UAV interaction systems is evolving from engineer-driven to user-driven, aiming to replace traditional predefined Human-UAV Interaction designs. This shift focuses on enabling more personalized task planning and design, thereby achieving a higher quality of interaction experience and greater flexibility, which can be used in many fileds, such as agriculture, aerial photography, logistics, and environmental monitoring. However, due to the lack of a common language between users and the UAVs, such interactions are often difficult to be achieved. The developments of Large Language Models possess the ability to understand nature languages and Robots' (UAVs') behaviors, marking the possibility of personalized Human-UAV Interaction. Recently, some HUI frameworks based on LLMs have been proposed, but they commonly suffer from difficulties in mixed task planning and execution, leading to low adaptability in complex scenarios. In this paper, we propose a novel dual-agent HUI framework. This framework constructs two independent LLM agents (a task planning agent, and an execution agent) and applies different Prompt Engineering to separately handle the understanding, planning, and execution of tasks. To verify the effectiveness and performance of the framework, we have built a task database covering four typical application scenarios of UAVs and quantified the performance of the HUI framework using three independent metrics. Meanwhile different LLM models are selected to control the UAVs with compared performance. Our user study experimental results demonstrate that the framework improves the smoothness of HUI and the flexibility of task execution in the tasks scenario we set up, effectively meeting users' personalized needs.}


\maketitle

\section{Introduction}\label{intro}

With the increasing popularity of Unmanned Aerial Vehicles (UAVs) in modern society, the complexity of Human-UAV Interaction (HUI) is also escalating. 
According to research by Rearch and Markets, the world's largest market research organization, the global UAVs market will reach a value of up to €53.1 billion by 2025~\cite{hohrova2023market}. The Civil Aviation Administration of China published~\cite{drone_report_2024} that the number of registered UAVs in China had reached 1.2627 million by the end of 2023, representing a 32.2\% increase from 2022. Clearly, the rapid proliferation of UAVs in human society has made the need to lower the threshold for HUI even more urgent. It is becoming more important that exploring how to control UAVs through diversified user interfaces and interaction mechanisms, and conducting in-depth research on more complex interaction designs from the perspectives of personalization, privacy protection, naturalness of interaction, security, and humanistic care~\cite{mirri2019human}. This trend underscores the increasingly refined nature of HUI.

Traditional control schemes are designed to meet the predefined, fixed HUI or task planning requirements set by engineers. For example, paper~\cite{kassab2020real,maher2017realtime} used neural networks to recognize human gestures for the purpose of enabling HUI. Rajappa et al~\cite{rajappa2017design} obtained human-transmitted signals by applying external forces to the drone and allowing its sensors to recognize the applied forces. In widely used agricultural UAVs, each task is executed based on task routes and parameters set by engineers, which are suitable for static scenarios. However, such control schemes have limited capabilities in handling complex tasks and often require human intervention to meet demands.
To address reliability and system manageability, classical robotics research has developed three-layer architectures~\cite{haddadin2016physical,de2008atlas,castrillo2022review}, which decompose capabilities hierarchically: a task-planning layer formulates goals, a behavior-control layer sequences predefined actions, and an execution layer interfaces with hardware. These architectures perform stably in domains such as UAV safety~\cite{castrillo2022review} and physical HUI~\cite{haddadin2016physical}. However, their task decomposition relies on hard-coded logic such as state machines, limiting their adaptability. For example, when a user requests ``avoid birds and inspect roof cracks," such architectures would typically require manually redefining state transitions and updating control graphs—an inflexible and labor-intensive process.

To improve modularity and reusability, skill-based architectures~\cite{rajappa2017design,dong2024bio} encapsulate capabilities into callable modules. These frameworks support task composition by assembling existing skills, such as Dong et al.’s marine disturbance controller~\cite{dong2024bio}. Nevertheless, they often demand retraining of parametric models to accommodate new skills, leading to scalability and generalization challenges.

Therefore, future HUI framework is evolving towards user-driven personalized task directions, aiming to provide high-quality HUI experiences and enhance the flexibility of task planning~\cite{obrenovic2024generative,apraiz2023evaluation}. Nevertheless, achieving user-driven personalized goals through traditional methods is challenging due to the lack of a common language between users and UAVs. This absence of a common language manifests in two aspects. Firstly, due to the complexity and abstraction of programming languages, humans often find it difficult to model and solve real-world problems using code, especially for non-programmers. Thus, even if users understand the essence of a problem, they may struggle to translate it into a format that UAVs or other automated framework can execute. Secondly, humans cannot directly communicate with algorithms, machines, or programs, as they operate based on logical rules and data inputs, rather than natural language or intuitive understanding. Consequently, users may not be able to convey specific instructions or requests in a way that UAVs can understand and respond to, which can lead to misunderstandings, errors, and ultimately, a suboptimal level of personalization and effectiveness in HUI.

\begin{figure}[H]
    \centering
    \includegraphics[width=0.8\linewidth]{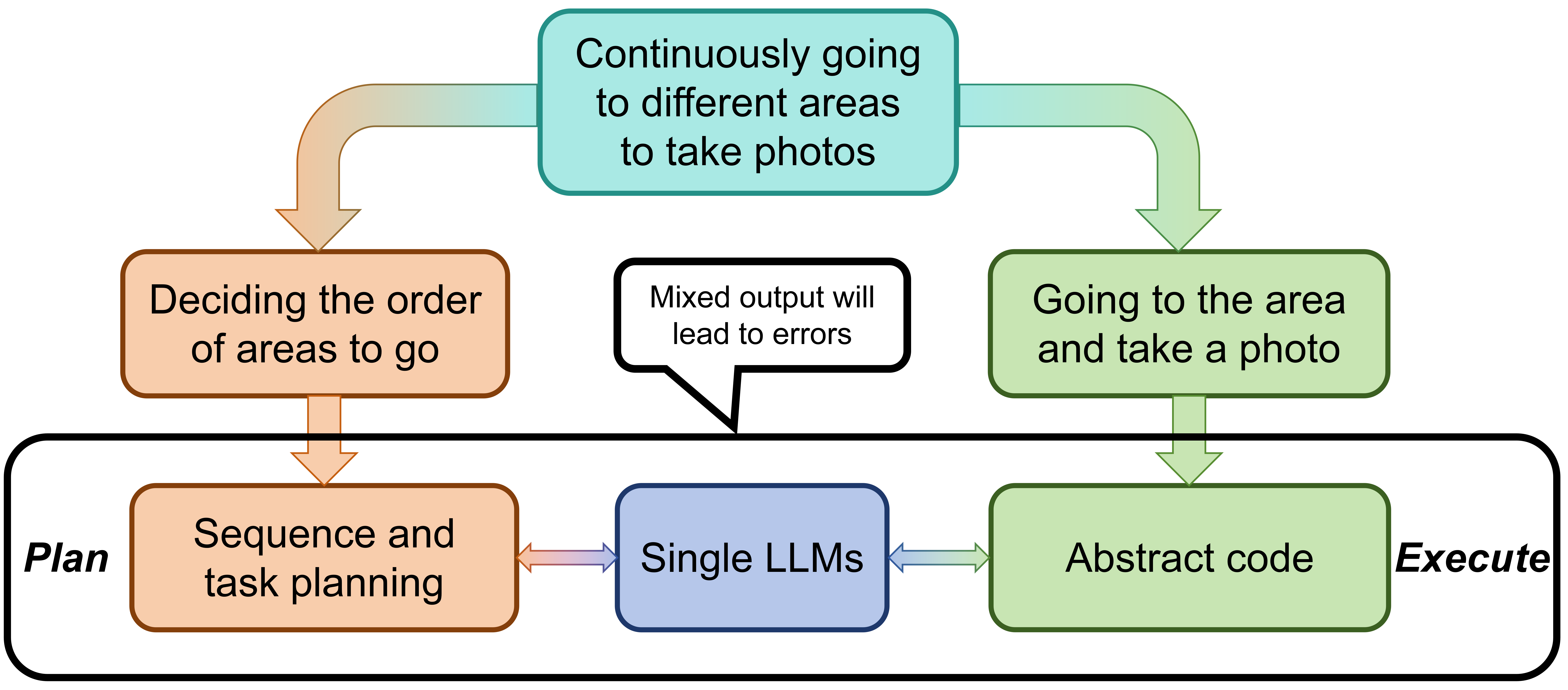}
    \caption{\label{fig:ques}Task decomposition divides a complex task into planning and execution phases, which differ for LLMs. Using one LLM-agent for both may cause errors, incorporate execution details into the plan, or conversely, insert planning elements into execution details.}
\end{figure}

Fortunately, with the emergence of ChatGPT, Large Language Models (LLMs) have developed rapidly, and the advent of Llama~\cite{touvron2023llama} has made it a reality for individuals to deploy LLMs locally, the emergence of LLMs points to a solution to provide a common language or a swift transform between human and machines. Trained on massive amounts of textual data, LLMs possess powerful semantic understanding capabilities, enabling them to deeply comprehend the syntax, semantics, and context of natural language. At the same time, the syntactic standards, logical consistency, and abstraction and modularization features of code allow LLMs to excel in understanding and generating code. These two characteristics make LLMs capable of serving as a direct bridge between users and UAVs, marking the possibility for users to interact directly with UAVs beyond the predefined settings of engineers. Recently, many HUI frameworks based on LLMs have been proposed to realize this vision: for example, using ChatGPT directly for task planning~\cite{singh2023progprompt} or integrating LLMs into the UAV control process through a Python pipeline~\cite{vemprala2024chatgpt}, leveraging LLMs to abstract code for users, thereby enabling them to organize personalized tasks more freely. Liu et al~\cite{liu2024llm} and Sun~\cite{sun2024generative} have integrated LLMs into the visual inspection of infrastructure in order to harness the ability of LLMs to comprehend human intentions and generate control commands.  However, these single-agent LLM methods still encounter planning failures when faced with complex planning challenges. The issues leading to such failures lie in:

\begin{enumerate}
    \item \textbf{Task Planning: }
    When single-agent LLM methods engage in task planning, they may erroneously integrate code or execution scripts into the plan, driven by the dual necessity of planning and execution, as illustrated in Figure~\ref{fig:ques}. This incorporation introduces errors, as the plan ideally should abstract from specific code or execution intricacies. In the context of task planning, LLMs are expected to concentrate on comprehending the task and its context, akin to a 'human' role. Conversely, during task execution, LLMs need to generate precise code or machine-readable instructions, functioning more like a 'script'. In practical scenarios, achieving a seamless transition between these 'human' and 'script' roles within a single LLM run presents a significant challenge. Consequently, when confronted with intricate planning and tasks, single LLM agent methods may encounter reduced planning efficiency or execution failures.
    \item \textbf{Task Execution: }Furthermore, when using LLMs for task planning or executing independent tasks, if the complexity of the task exceeds the capabilities of the predefined high-level function library, the existing HUI frameworks based on LLMs~\cite{vemprala2024chatgpt} may fail to execute the task, especially for complex operations like obstacle avoidance. For UAVs, obstacle avoidance involves an intricate process, including receiving camera data, inferring surrounding obstacles, and calculating subsequent velocity vectors, which cannot be simplified into a script. Therefore, in such cases, the interaction framework must possess the ability to invoke tools to complete the task. Similarly, for task planning, when the complexity of the task surpasses the solo handling capacity of LLMs, the framework should also have the capability to call upon tools.
\end{enumerate}

To address the aforementioned issues, we propose ``UAV-GPT," a dual-agent framework tailored for UAV interaction. It integrates the natural language understanding capabilities of LLMs via API interfaces, enabling UAVs to ``understand" human language and learn to invoke tools to tackle complex task planning and execution challenges. The framework consists of two LLM agents: a planning agent and an execution agent. To solve the task planning problem, the planning agent utilizes the natural language understanding abilities of LLMs to classify and plan tasks, accurately categorizing user needs and formulating reasonable execution plans through a predefined behavior library and discrete task reorganization. It then conveys the classified and planned tasks to the execution agent. To solve the task execution problem, execution agent will select the most suitable method and executes the corresponding operational code based on the mapping relationship between a predefined command library and the LLM, ensuring precise and efficient execution of every instruction. To enhance generalization, we integrated ROS-based control algorithms which fall within the scope of the skill base architecture, the difference is that we use skills as a supplement to the robotics framework rather than as its foundation. Meanwhile, within the classical three-layer architecture, our planning and execution agents operate within the scope of the task planning and execution layers, respectively. Crucially, the integration of LLMs significantly optimizes these layers compared to traditional implementations: it empowers the planning layer with semantic reasoning to transcend hard-coded logic, and upgrades the execution layer with code comprehension to enable dynamic rather than static tool invocation. This ensures that, unlike purely skill-based or traditional architectures, our system achieves better flexibility and adaptability.

In summary, our contributions are as follows:

\begin{itemize}
      \item We propose a dual-agent intelligent interaction framework based on LLMs for UAV, which achieves precise classification of tasks and outputs solutions with high success rates and efficiency. Planning agent employs a predefined behavior library and discrete task reorganization to plan complex tasks reasonably, execution agent convert the solutions into precise machine language outputs.
      \item We introduced traditional control algorithms to the execution agent to enhance its range of applicability. The execution agent evaluates tasks through the mapping relationship between a predefined command library and LLMs, and then selects a reasonable execution method to complete the tasks.
      \item The user study demostrate that our framework improves HUI smoothness and task-execution flexibility, enabling better support for personalized user needs.
      \item The simulation and real experiments results demonstrate that, compared to the HUI framework with a single LLM agent, our framework achieves an average improvement of 60\% in operational efficiency for complex tasks and a 30\% increase in task execution success rate.
      
\end{itemize}

In the overview of Section~\ref{me}, we first briefly explained the limitations of the traditional HUI for UAVs, then describe the specific architectural approach of UAV-GPT for the HUI scheme, where the interaction requirements are obtained from users through simulated requests.
In Section~\ref{3.1}, we construct a database related to daily interaction tasks with UAVs. We have defined four task categories for the database and provided detailed explanations for the classification rationale. Within each task category, we have designed 20 different tasks that encompass simple movements to continuous complex route planning and optimal solutions. Simultaneously, we quantify the performance of different HUI frameworks on this database using three indicators, which allows for a clearer demonstration of the advantages of our framework.
In Sections~\ref{3.2} and~\ref{3.3}, we provide detailed explanations of the methods at both agents of the interaction framework. The overall architecture is primarily implemented using ERNIE 4.0 and evaluated through performance analysis.

\section{Related work}

\subsection{Traditional Human-Robot Interaction Methods}

\textbf{Physical Human-Robot interaction (PHRI)} refers to direct and intuitive communication between humans and robots without intermediary mediums, effectively conveying rich tactile feedback to humans. In the early stages of robotics, due to the lack of tactile sensing capabilities, traditional interaction methods relied on wearable devices. For example, Kim et al~\cite{kim2005force} developed an exoskeleton master arm capable of detecting torque applied by users and providing multimodal contact feedback, enabling device mobility. Another example is the 
vibrating motor-controlled bracelet proposed in~\cite{scheggi2012vibrotactile}, 
which guides robots to follow human trajectories. Haddadin et al~\cite{haddadin2016physical} summarized the latest advancements in PHRI, while De Santis et al~\cite{de2008atlas}  focused on addressing the safety aspects of PHRI.

However, these interaction methods face a common limitation: the interaction distance is typically fixed and confined to a small range. Moreover, the forms of robots capable of interaction are limited, restricting effective interaction with non-anthropomorphic robots such as UAVs. To address this challenge, researchers have proposed ``Teleoperation Human-Robot Interaction (THRI)." In the following, we provide an overview of this area.

\noindent\textbf{Teleoperation Human-Robot Interaction} 
leverages the concept of indirect control, where commands are transmitted through wireless communication channels. This significantly enhances the effective distance range in human-machine interaction. For example, Peppoloni et al~\cite{peppoloni2015immersive} designed a ROS-integrated interface that enables users to remotely control robots via hand gestures. Similarly, Tsetserukou et al~\cite{tsetserukou2007towards} proposed a method for teleoperating robotic arms using full-body motion. Su et al~\cite{su2022mixed} improved traditional teleoperation by combining 2D image transmission with a 3D virtual interface, enhancing spatial awareness during remote control. Fani et al~\cite{fani2018simplifying} explored wearable-based interaction, offering a simpler and more intuitive experience that improves responsiveness.

These studies have greatly extended the restricted interaction distances and changed the forms of robotic agents at the execution end in physical human-robot interaction. However, they also introduced new challenges. When a user engages in remote control of a robot, it is imperative that the robot operates within a size or scale that is commensurate with the user's environment, or adheres to predefined system scales. However, the inherent variability in environmental dimensions precludes an exact correspondence, thereby necessitating user reliance on these predetermined scales. This reliance, in turn, imposes limitations on the efficacy of remote control in practical, real-world scenarios. Hence, a pertinent question arises: Can the integration of advanced technologies, specifically LLM-based Human-Robot Interaction methods, empower the execution end (i.e., the robot) with the capacity to interpret and execute commands in a manner that facilitates more personalized responses tailored to user-specific requirements? This exploration aims to transcend the limitations imposed by reliance on predefined scales and enhance the effectiveness of remote control in real-world applications.

\noindent \textbf{Supervisory and Interactive Systems in UAV Operations:}
\noindent Supervisory control architectures enable strategic human oversight in UAV operations through dynamic task reconfiguration and hierarchical monitoring. Wang et al~\cite{lahmeri2024robust} mitigate adversarial environmental uncertainties by formalizing swarm task planning as a Constrained Markov Decision Process (CMDP). Their performance-function-guided SAC-Lagrangian algorithm incorporates safety boundaries (e.g., no-fly zones) into optimization objectives, achieving 96\% mission success rates under electronic warfare conditions. Dong et al~\cite{dong2024bio} developed a bio-inspired sliding mode controller with radial basis function neural networks (RBFNN) for marine rescue drones, using Multi-Layer Perceptron (MLP) to compensate for ocean disturbances and thruster faults. This approach ensures trajectory stability under wind and wave perturbations while reducing computational complexity, validated through Lyapunov stability proofs.

As for the ground station design, standardization and modularity are critical for scalable UAV command interfaces. KP Arnold et al~\cite{arnold2016uav} primarily examine the types, components, safety features, redundancy design, and future applications of UAV Ground Control Stations (GCS). E Çintaş et al~\cite{ccintacs2020vision} primarily examine how to achieve visual tracking of moving targets using UAVs equipped with low-cost hardware and novel GCS. The core objective of this system is to develop a visual tracking system capable of operating efficiently in real-time applications, particularly within complex environments featuring non-fixed perspectives and dynamic target motion.

Mission planning integrates path optimization with threat mitigation optimization with threat mitigation. Xiong et al~\cite{xiong2023multi} combined adaptive genetic algorithms (AGA) and sine-cosine particle swarm optimization (SCPSO) for multi-drone disaster rescue, dynamically assigning tasks while generating collision-free 3D paths around collapsed structures. Security frameworks must counter emerging threats: Castrillo et al~\cite{castrillo2022review} established layered protocols including MAVLink encryption, RF signal fingerprinting to detect rogue drones, and GPS spoofing countermeasures deployed in critical infrastructure protection. This dual focus on operational efficiency and electronic security addresses gaps in remote UAV supervision.

\subsection{Classical Robotic Methods}
\textbf{Classical robotics architectures} (e.g., Three-Layer ~\cite{haddadin2016physical,de2008atlas,castrillo2022review}) Classical robotics architectures, such as the Three-Layer Architecture~\cite{haddadin2016physical,de2008atlas,castrillo2022review}, ensure system reliability by structuring tasks into hierarchical layers. The task-planning layer decomposes complex goals into simpler sub-tasks, the behavior-control layer schedules predefined actions, and the execution layer controls hardware interfaces. These architectures have proven effective in domains like UAV security~\cite{castrillo2022review} and physical human-robot interaction~\cite{haddadin2016physical}, where reliability and stability are paramount. However, they typically rely on predefined state machines or fixed task decompositions, which can be limiting when responding to more dynamic user instructions. For example, when users provide instructions like ``avoid birds and inspect roof cracks," traditional three-layer systems require manual redesign of state transitions to incorporate new tasks.

\noindent \textbf{Skill-based architectures}~\cite{rajappa2017design,dong2024bio} modularize capabilities by encapsulating them into reusable skill modules, making the system more flexible and adaptable. These frameworks allow for easy composition of skills to perform various tasks. For example, Dong et al~\cite{dong2024bio} developed a marine disturbance controller that can be used to address specific environmental challenges. However, extending these frameworks with new skills often requires retraining or reconfiguring parametric models, which can be resource-intensive.

Relative to these frameworks, our planning and execution agents map to the task-planning and execution layers, respectively, with ROS based control algorithms serving as a supplement to the execution agent. The integration of LLMs improves interaction flexibility and generalization compared to these frameworks.

\subsection{LLM-based Human-Robot Interaction methods}

Dialogue, as a natural mode of human communication, contributes to integrating robots into human society in the context of HRI. The emergence of LLMs with advanced natural language understanding capabilities enables them to effectively assist robots in understanding scale differences in various environments. They also facilitate the integration and transmission of tasks from the control side to the execution side (robots).

\noindent\textbf{Human-Robot}: Wang et al~\cite{wang2024llm} directly utilized LLMs to generate action commands, but significant deviations existed. Mai et al~\cite{mai2023llm} using Large-scale Language Model as a robotic brain to unify egocentric memory and control. Mower et al~\cite{mower2024ros} extracted actions from the output of LLM and executed ROS operations/services. Ishika Singh et al in~\cite{singh2023progprompt} proposed a code framework for task planning using ChatGPT. The core of these articles are to provide program specifications of available actions and objects in the environment to the LLMs, which then plan all possible actions. Their experiment confirmed the feasibility of this approach. Since the task processes provided by LLMs are not classified, they can only transmit simple, encapsulated low-level control functions to LLMs, inform them of specific scenarios and tasks that need to be executed, and then let LLMs process the task flows and issue commands to the execution end.

In contrast, Anis Koubaa \sout{et al.} chose to integrate LLMs with ROS2. In~\cite{koubaa2023rosgpt}, they explored the method of integrating ChatGPT into ROS2 and successfully designed a ROS2 package that could convert human language instructions into navigation commands for ROS2 robots. This package-based adaptation method can be more universally applied to other robot projects, thus having higher compatibility. However, their method only differs from~\cite{singh2023progprompt} in terms of the path implementation. The classification of tasks and the execution of complex tasks have not been perfectly addressed.

\noindent\textbf{Human-UAV}: Regarding the interaction between UAVs and LLMs, the challenges we face are similar to those discussed in the previous two articles. Javaid et al~\cite{javaid2024large} have investigated the feasibility of combining UAVs with LLMs. With the rapid development of LLMs. Sai Vemprala et al~\cite{vemprala2024chatgpt} utilized ChatGPT to achieve UAV control and path planning. Regarding the issue of  UAV mission planning, they also demonstrated simple autonomous obstacle avoidance behaviors controlled by ChatGPT in AirSim simulations. Their work demonstrated the efficiency of using the ChatGPT language model for robot control. Cui et al~\cite{cui2024tpml} employed LLMs to plan UAV missions, utilizing a single natural language input to command multiple UAVs in both synchronous and asynchronous modes.
Lykov et al~\cite{lykov2024flockgpt} have achieved rapid swarm control of  UAVs utilizing GPT. The described method enables intuitive orchestration of swarms of any size to achieve desired geometric shapes.

However, since they only used the UAV SDK control function (i.e., controlled by Python scripts) without integrating the ROS platform, and they only used a single LLM-agent to simultaneously handle task planning and execution, using pre-prompts for function learning in LLMs without separating planning and execution, their architecture could not accurately implement the planning and execution of long and complex tasks. Mohamed Lamine Tazir in reference~\cite{tazir2023words}, improved the hardware equipment and used ChatGPT to transmit control information to UAVs based on PX4/Gazebo simulations, successfully controlling the UAV's operations. This was a very promising attempt because it combined ROS and ChatGPT in their system, but issues with task classification and long-process planning still persist. If we want LLMs to demonstrate their assistive capabilities in more scenarios, such as obstacle avoidance, path planning, or longer task planning, we need multiple LLM-agents and a deeper integration of ChatGPT into ROS nodes.

\begin{table}[h]
    \caption{Comparison of Methods: Here, we clearly demonstrate the advantages of our method compared to traditional physical and teleoperation HUI, as well as single-agent HUI based on LLM.}\label{tab1}%
    \begin{tabular}{@{}>{\centering\arraybackslash}m{1.5cm}|
        >{\centering\arraybackslash}m{1.2cm}|
        >{\centering\arraybackslash}m{1.4cm}|
        >{\centering\arraybackslash}m{1.4cm}|
        >{\centering\arraybackslash}m{1.2cm}|
        >{\centering\arraybackslash}m{1.2cm}|
        >{\centering\arraybackslash}m{1.2cm}|
        >{\centering\arraybackslash}m{0.8cm}@{}}
    \toprule
    HUI & Force Reflected ~\cite{kim2005force} & ROS-Integrated Framework~\cite{peppoloni2015immersive} & Prog-Prompt~\cite{singh2023progprompt} & ROS-LLM~\cite{koubaa2023rosgpt, mower2024ros} & Prompt Craft~\cite{vemprala2024chatgpt} & Words to Flight~\cite{tazir2023words} & \textbf{Ours} \\
    \midrule
    Remote Control & \ding{55} & \checkmark & \checkmark & \checkmark & \checkmark & \checkmark &\checkmark\\
    \midrule
    Personalized Tasks & \ding{55} & \ding{55} & \checkmark & \checkmark & \checkmark & \checkmark&\checkmark \\
    \midrule
    High-Level Plan & \ding{55} & \ding{55} & \ding{55} & \ding{55} & \ding{55} & \ding{55} & \checkmark \\
    \midrule
    Tools Ability & \ding{55} & \ding{55} & \ding{55} & \ding{55} & \ding{55} & \ding{55}&\checkmark \\
    \botrule
    \end{tabular}
\end{table}

\section{Methods}
\label{me}
\subsection*{Overview}
\label{Ov}

\begin{figure}[H]
    \centering
    \includegraphics[width=1\linewidth]{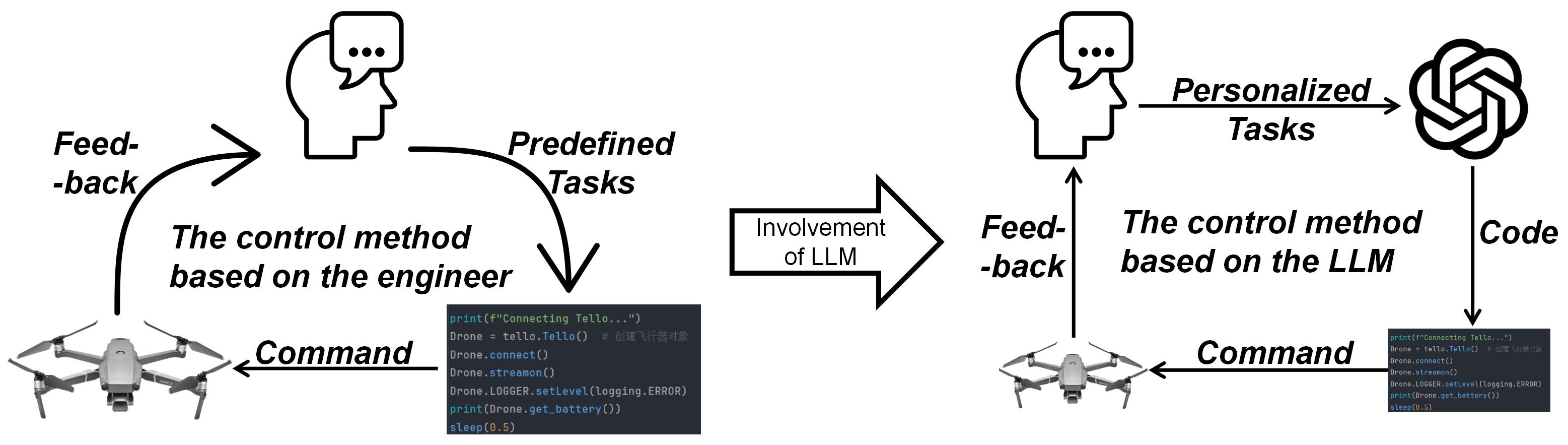}
    \caption{\label{fig:traditional}Traditional teleoperation control method}
\end{figure}

The traditional method for HUI with UAVs is depicted in Figure~\ref{fig:traditional}. It primarily hinges on users' ability to abstract real-world tasks into coding representations and subsequently transmit these to the UAV via wireless links. Inherently, this operational procedure presents an obstacle, as not all users own the capability to translate real-world tasks into code. This is precisely why it is necessary to introduce LLMs into this process. To more concisely and clearly describe UAV-GPT, we characterize it as a converter USR2ML  based on LLMs that takes users' speech/text as input and converts it into executable machine language vectors (MLV):
\begin{equation}
    l=(l_1,l_2,l_3...,l_T)
\end{equation}
Where $l$ denotes the length of the Machine Language Vector (MLV). For each time step $t \in {1,...,T}$, the output length $l_t$ should be kept within a reasonable range, as the UAV is required to execute the complete instruction in a single step. If the output is too long, it may result in interaction failure due to system overload or motion deviation. In particular, when the execution length exceeds a certain threshold, the UAV's trajectory may exhibit significant drift caused by cumulative errors from onboard sensors and flight controllers. Since such drift is closely tied to hardware limitations—which are beyond the scope of this study—we define a reasonable range of execution length through three points to minimize the impact of hardware-induced errors and ensure task reliability. This reasonable range is formally defined as a constraint ensuring UAVs execute complete MLVs without exceeding the system’s physical or computational capacity. This constraint is determined by: 
\begin{itemize}
    \item Processor limitations (Tello SDK buffer capacity)
    \item Real-time communication stability (WiFi packet size thresholds)
    \item Safety protocols (max command chain before failsafe activation)
\end{itemize}
Experimentally, we set $l_{\min}=3$ (e.g., takeoff$\rightarrow$hover$\rightarrow$land) and $l_{\max}=7$.

When UAV-GPT outputs MLV (Machine Language Vector), we will detect its length. For outputs with a length exceeding $l_{\max}$, we consider the system's task execution as failed. The reason is that the task language received by the UAV should be within a specified range; otherwise, it may fail to execute due to hardware resource constraints. Meanwhile, for tasks with longer processes, one of the design goals of the planner is to split the overall task into multiple sub-segments that can be executed in a single run. Therefore, if the output length exceeds $l_{\max}$, it can be regarded as a failure of the planner in task division.

To achieve users' requirements, UAV-GPT first needs to classify the user's task. The application scenarios of UAVs in reality are complex and variable, and identifying the accurate task type will significantly enhance the success rate and efficiency of the agent. Next, it needs to solve the problem and output a machine language vector that meets the requirements. To achieve this goal, we propose a dual-agent architecture, as shown in Figure~\ref{fig:3}. The first stage of UAV-GPT is a planning agent, whose task is to convert user intentions into fixed categories of tasks and reorganize the optimal solutions through a predefined behavior library and discrete task reorganization. The second stage is a machine language execution agent, whose task is to evaluate the task flow and output machine language within a reasonable range by leveraging the mapping between the LLM and the predefined instruction library.

\subsection{Task Classification}
\label{3.1}

In the task classification framework, we adopt a two-dimensional tagging system as shown in Table~\ref{tab2}, where the first core dimension is the \textbf{``Simple-Complex"} continuum. This dimension uses the following two sub-dimensions to characterize:
\begin{itemize}
    \item \textbf{State Space Complexity $S_c$}: It is jointly determined by the number of monitoring points $p \in C_t$ and the range of dangerous areas $d\in A$, where the $C_t$ is the content of task and $A$ is the space of task in System Prompt(which is the preset system knowledge of the current task scene), reflecting the amount of perceptual information that needs to be focused on for the task.
    \item \textbf{Motion Space Complexity $M_c$}: It is determined by the length $l \in C_t$ of the action sequence, reflecting the reasoning depth required for the large language model to solve the problem.
\end{itemize}

In human-computer interaction scenarios, an increase in the number of task monitoring points $p$ and the range of dangerous areas $d$ means more optional paths and execution conditions, which directly exacerbates the complexity of the task~\cite{zhu2023complexity}. Meanwhile, cognitive load theory~\cite{sweller2011cognitive} indicates that a larger number of monitoring points and an expanded range of dangerous areas will cause LLMs to process richer context of task $C_t$ and perform more intermediate reasoning steps before executing the task, ultimately increasing the complexity of the task.

\begin{figure}[H]
    \centering
    \includegraphics[width=0.6\linewidth]{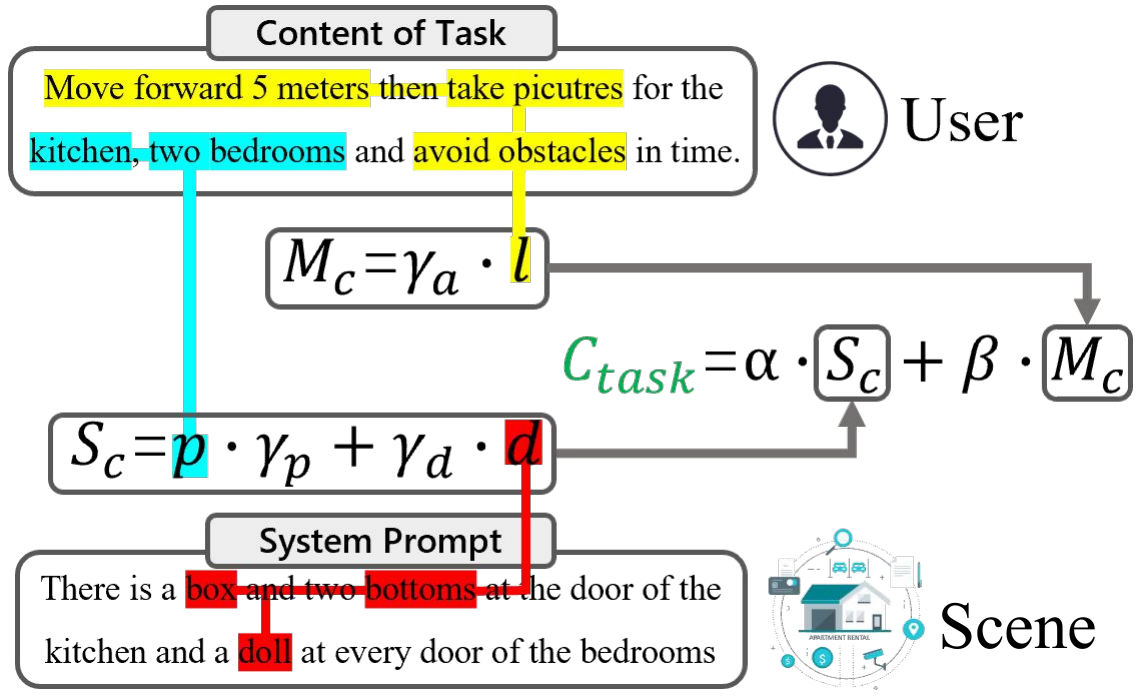}
    \caption{\label{equpicture}The red parts represent dangerous areas, the yellow parts represent action sequences, and the blue parts represent monitoring points.}
\end{figure}

\begin{figure}[H]
    \centering
    \includegraphics[width=1\linewidth]{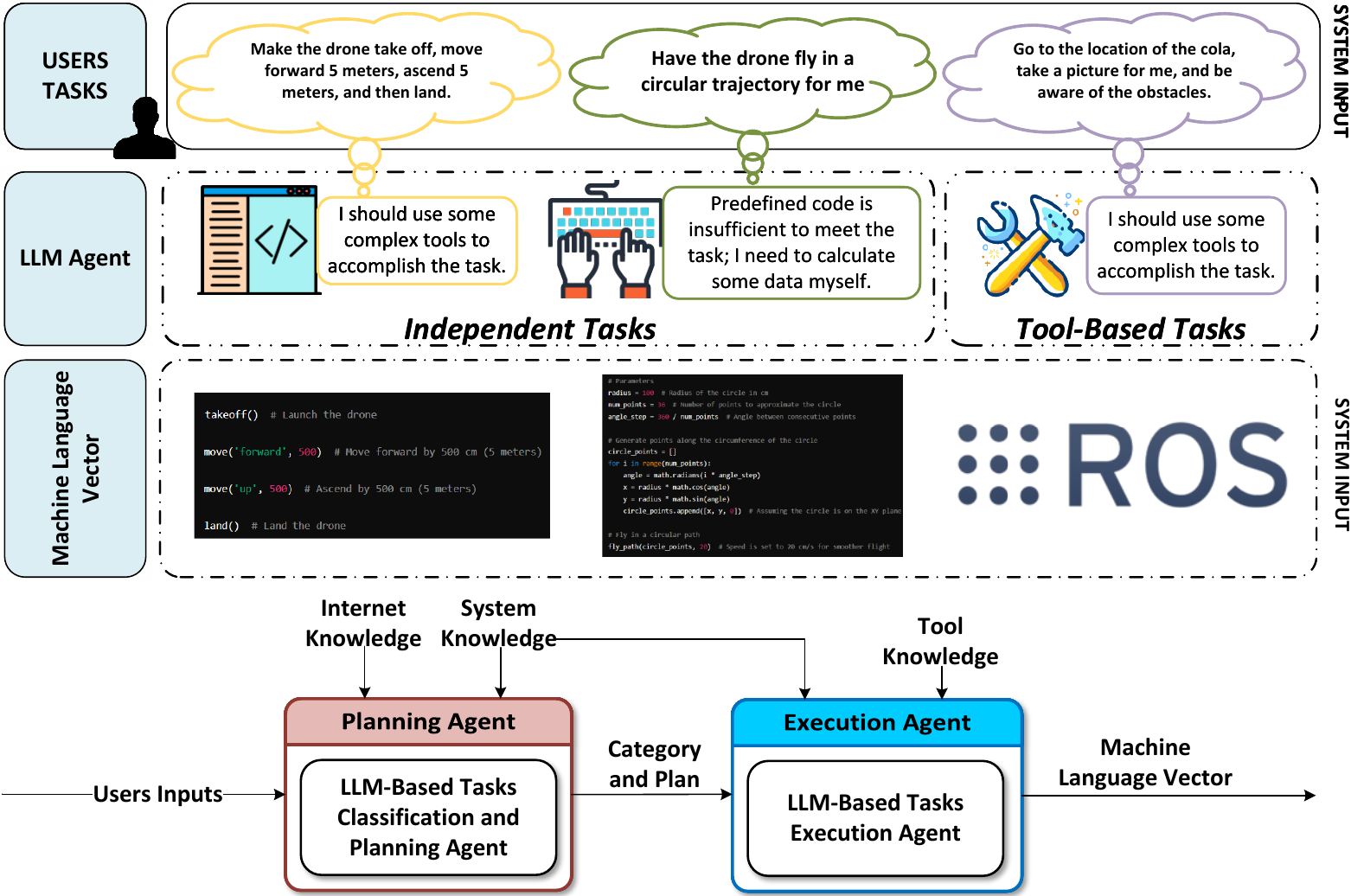}
    \caption{\label{fig:3}This is the dual-agent architecture for user's requests to machine language vector.}
\end{figure}

Next, we define the overall task complexity as a linear combination:
\begin{equation}
    \label{equ:TASK}
    \mathcal{C}_{\text{task}} = \alpha \cdot S_c + \beta \cdot M_c,
\end{equation}
here, $S_c = \gamma_p \cdot p + \gamma_d \cdot d$, which is the state space complexity determined by the number of monitoring points $p$ and the range of dangerous areas $d$. $M_c = \gamma_a \cdot l$, which is the motion space complexity determined by the action sequence length $l$ required by the task. Figure~\ref{equpicture} shows how we decide these values during a task. 

Note that our tasks are strictly limited to a 50×50×50 meters volume. This prevents conflicts where simple tasks requiring long-distance execution receive an inaccurately low $\mathcal{C}_{task}$ relative to their actual complexity. Instructions must also be explicit and non-autonomous to ensure that $p$ and $q$ remain computable, even when their values are zero. For example, high-autonomy tasks ``search for a specific location” are excluded due to the infeasibility of defining $p$. In contrast, commands ``take off and move forward 5 meters” remain valid for setting $p$ and $q$ to zero simply shifts the final complexity $\mathcal{C}_{task}$ to action sequence length $l$ without compromising classification accuracy.

The coefficients $\alpha$ and $\beta$ control the balance between perception and reasoning burdens, while $\gamma_p$, $\gamma_d$, and $\gamma_a$ serve as unit scaling factors.
Based on this score, we define a two-level classification rule:
\begin{equation}
\text{Task Type} =
\begin{cases}
\text{Simple}, & \text{if } \mathcal{C}_{\text{task}} \leq \theta \\
\text{Complex}, & \text{if } \mathcal{C}_{\text{task}} > \theta
\end{cases}
\end{equation}
the threshold $\theta$ is an empirical or learned parameter that reflects the separation boundary in the “simple-complex” continuum.
To determine the coefficients $\gamma_p$,$\gamma_d$,$\gamma_l$ and $\theta$ involved in the task complexity Formulation~\ref{equ:TASK}, we utilize two expert-annotated datasets with high/low complexity task labels: CLAD~\cite{verwimp2023clad} and BRMData~\cite{zhang2024empowering}. By analyzing the $p$, $d$, and $l$ within these datasets and correlating them with the provided task complexity labels, we estimate the relative weights of each factor. In this work, we empirically set the balance coefficients as $\alpha = \beta = 0.5$, assuming equal contributions from state space complexity and motion space complexity for simplicity.

For the second dimension of Table~\ref{tab2}, \textbf{``Independent vs. Tool-assisted"}, we adopt a knowledge-based decision mechanism. Specifically, the system incorporates two types of prior knowledge:
\begin{itemize}
    \item{\textbf{Internet Knowledge}} refers to general, preloaded knowledge that forms the foundational information base of the robot;
    \item{\textbf{System Knowledge}} refers to task-specific knowledge dynamically generated or loaded for the current interaction, including contextual and environmental information relevant to the scenario.
\end{itemize}
\noindent During execution, we extract \textbf{task-related keywords} $K$ from the input instruction $C_t$: $K=\{k_1, k_2,...,k_n\}$ where $k_i \in \mathrm{Keywords}(C_t)$, and match them against the keywords present in system knowledge and Internet knowledge: $K_{total}=K_{sys} \cup K_{net}$. If the extracted keywords can be fully matched with the system knowledge, the task is classified as Independent, as it can be handled with existing system information: $\forall k \in K, k \in K_{total} \Rightarrow \mathrm{Independent}$. Otherwise, if the keywords require additional unknown tools or capabilities beyond the knowledge bases, the task is labeled as Tool-Assisted, implying dependence on external modules or resources: $\exists k \in K, k \not\in K_{total} \Rightarrow \mathrm{Tool-Assisted}$.

This classification approach allows the system to dynamically determine task autonomy based on the semantic alignment between user instructions and available knowledge, thereby bridging high-level language understanding with execution-level resource requirements. The generalizability of the proposed framework is currently bounded by the calibration dataset used during the coefficient learning phase. Specifically, the classification of UAV interaction tasks relies on the coefficient ranges derived from this dataset. Therefore, the broader and more diverse the calibration data, the wider the range of tasks the system can effectively support. This design ensures a balance between interpretability and scalability, allowing the framework to gradually extend its capability as more representative task data becomes available. In our implementation, we have made efforts to enrich the calibration dataset and ensure that the resulting classification coefficients are both reasonable and robust. By combining these two dimensions, we derive four distinct task types: \textbf{SI,ST,CI,CT}.

\begin{table}[h]
    \caption{Classification Tags}\label{tab2}%
    \begin{tabular}{@{}>{\centering\arraybackslash}m{2cm}|
        >{\centering\arraybackslash}m{4cm}|
        >{\centering\arraybackslash}m{5cm}@{}}
    \toprule
     & Simple  & Complex\\
    \midrule
    Independent    & Move forward 5 meters and take a picture   
                   & Move forward 5 meters then take pictures for kitchen and two bedrooms\\\hline
    Tool-assisted & Move forward 5 meters and avoid obstacles in time
                  & Move forward 5 meters then take pictures for the kitchen and two bedrooms and avoid obstacles in time\\
    \botrule
    \end{tabular}
\end{table}

\textcolor{red}{}

The planning agent of the UAV-GPT framework is responsible for comprehending user intent and accurately classifying tasks into the first dimension. The execution agent, meanwhile, need to decide if the task is independent, and choose the right method to do the task through the 2-dimensional classification, translates these classified tasks and plans into machine language for ultimate execution. In designing the framework, we leverage three key capabilities of LLMs: Firstly, converting real-world task descriptions into code tasks. Secondly, identifying task types. Thirdly, assessing tasks and selecting optimal implementation approaches through LLM's reasoning abilities.
To evaluate the performance of the two agents within the UAV-GPT framework, we employed three performance metrics: Intent Recognition Accuracy (IRA), Task Execution Success Rate (ESR), and UAV Energy Consumption (UEC) for qualitative testing.

\subsection{The design of Planning Agent}
\label{3.2}
The core function of the LLM-based task planning agent is to receive user voice or text instructions and perform complexity classification and planning accordingly Figure~\ref{fig:3}. To support this process, a two-dimensional task categorization system is introduced into the LLM, along with clearly defined classification rules. This enables the agent to interpret diverse natural language expressions of the same task and map them to appropriate structured plans. Specifically, the four task types include: simple independent tasks (e.g., “move forward 10 meters and then ascend”), simple tool-assisted tasks (e.g., “move forward 10 meters while avoiding obstacles in real-time”), complex independent tasks (e.g., “go to the kitchen, bathroom, and living room to take photos and return”), and complex tool-assisted tasks (e.g., “go to the kitchen, bathroom, and living room to take photos and return, avoiding obstacles in real-time during the journey”).

To enhance the recognition capability of the interaction framework, we have designed a specific guidance path for the planning agent. This path requires the LLM to extract key information from user inputs: Scene Keywords and Task Actions. The LLM infers the possibility of task planning scenarios based on Scene Keywords and differentiates execution methods through Task Actions. These two pieces of information correspond to the first and second dimensions of task classification, respectively. This guidance scheme enables the LLM to classify the two indicators separately, significantly improving classification accuracy.

Regarding task planning, there are notable differences between the planning and execution agents. The planning agent addresses the one-dimensional range issue in task classification, employing distinct planning approaches for tasks of varying complexities. For simple tasks, the LLM describes the action sequence the UAV needs to perform in precise natural language. For complex tasks, the LLM utilizes tools such as pre-trained neural networks or fine-tuned information to plan for optimal success rates and low energy consumption. In summary, the design of the planning agent encompasses selecting an LLM model (in this case, qianwen), utilizing prompts to design the guidance structure, and creating a carefully crafted 2D task classification list with well-structured context. This setup enables the planning agent to effectively accomplish classification and planning tasks within the UAV-GPT framework. To evaluate the performance of the planning agent in terms of IRA and UEC , we have constructed a UAV-GPT database for experimental testing.

\subsection{The Design of Execution Agent}
\label{3.3}
A common approach to tackling task execution for the UAV is to rely solely on an LLM agent, which utilizes LLMs such as qianwen or LLama3 to convert naturally described task workflows into machine language (code) comprehensible by the UAV, subsequently transmitting it via remote communication facilities. For instance, the CaP Framework~\cite{liang2023code} adopts this method. CaP is a robot-centric language model, generates programmatic representations capable of yielding reactive policies (like impedance controllers) and waypoint-based strategies (e.g., vision-based grasping and placing, trajectory-based control). The core of the CaP approach lies in linking classical logical structures, incorporating third-party libraries (e.g., NumPy, Shapely) for arithmetic operations, and allowing LLMs to receive commands and autonomously recombine API calls to produce novel strategy code. While promising, this method is constrained by its execution format, generating script-based strategies for robots to execute, limiting real-time interaction with robots during task execution and thus insufficient for tasks like real-time obstacle avoidance.

\begin{center}
\begin{algorithm}
\caption{\label{code}Main Loop for UAV-GPT, $p$ is monitoring points, $d$ is range of dangerous areas and $l$ is the length of action sequence in user's input}
\begin{multicols}{2}
\begin{algorithmic}[1]
\State \textbf{Initialize} Planner Agent
\State \textbf{Load} AirSim basic and system prompt
\While{true}
    \State \texttt{input} $\gets$ \Call{Get\{User\&Pre\}Input}{}
    \If{$\texttt{input}=\text{!quit}$ or $\texttt{input}=\text{!exit}$}
        \State \Call{ExitProgram}{}
    \ElsIf{$\texttt{input} =\text{!clear}$}
        \State \Call{ClearScreen}{}
    \Else
        \State $(p, d, l)$ = $LLM_{plan}$($\texttt{input}$)
        \State get $\texttt{complexity}$ {$\mathcal{C}_{\text{task}}$} based on Eq~\ref{equ:TASK}
        \If{$\mathcal{C}_{\text{task}}>\theta$}
            \State $\texttt{taskType1}=\texttt{Complex}$
        \Else
            \State $\texttt{taskType1}=\texttt{Simple}$
        \EndIf
        \State $\texttt{Plan}$ =$LLM_{plan}$(${C}_{t}$, $\texttt{taskType1}$)
        \State \Call{OutputPlan}{$\texttt{Plan}$}
    \EndIf
\EndWhile\\
\State \textbf{Initialize} Execution Agent
\State \textbf{Load} Tool basic and system prompt
\While{waiting for \textbf{Task Type1} and \textbf{Plan}}
\State \texttt{input1} $\gets$ \Call{Sys-Konwledge}{}
\State \texttt{input2} $\gets$ \Call{In-Konwledge}{}
\State $\texttt{keywords}$ = $LLM_{execut}$($C_t$)
\State \texttt{Pre\_K} = \texttt{input1}+\texttt{input2}
    \If{\texttt{keywords}==\texttt{Pre\_k}}
        \State \texttt{taskType2} = \texttt{Independent}
    \Else
        \State \texttt{taskType2} = \texttt{Tool\_Based}
    \EndIf 
    \If{$\texttt{TaskType2} = \texttt{Independent}$} 
        \State \texttt{action} = $LLM_{execut}$(\texttt{Plan})
        \State \Call{ExecutePlan} {LLM}
    \Else
        \State \texttt{action} = $LLM_{execut}$(\texttt{Plan})
        \State \Call{ExecutePlan} {LLM and Tool}
    \EndIf
\EndWhile
\end{algorithmic}
\end{multicols} 
\end{algorithm}
\end{center}

\begin{figure}[H]
    \centering
    \includegraphics[width=0.9\linewidth]{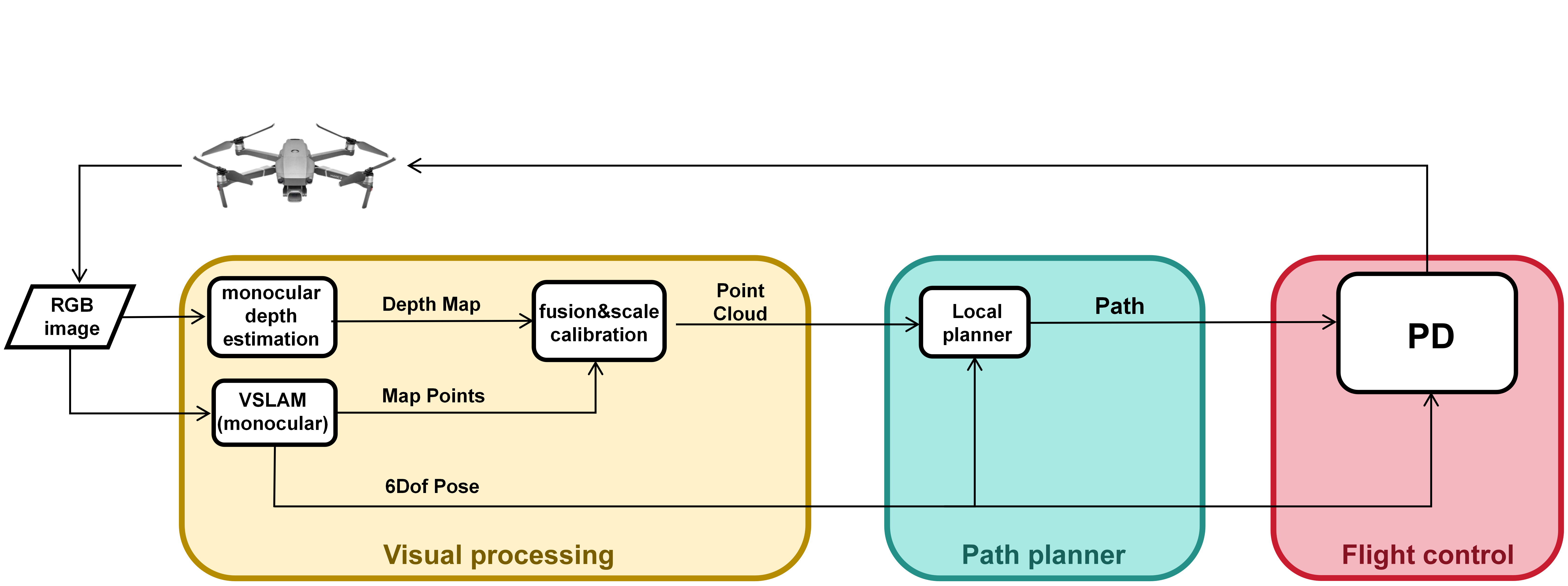}
    \caption{\label{fig:ROS}The UAV uses ROS with multiple algorithms to perform tasks: capturing an RGB image, processing it for depth estimation and 3D mapping, planning a path, and adjusting flight with a controller.}
\end{figure}

Therefore, building upon our execution agent architecture, we preserved the LLM's capability to translate natural language into machine language while introducing tool invocation abilities. As illustrated in Algorithm~\ref{code}, after the user provides an instruction, the system first computes the task complexity based on monitoring points $p$, danger range $d$ and input length $l$. The Planner Agent then will determine which kind of task type it is and generates a corresponding action plan. For the the Execution Agent, it is equipped with a neural network-based RGB image recognition framework integrated with egoplanner for monocular real-time obstacle avoidance (i.e., path planning functionality) in Figure~\ref{fig:ROS} and it will decide how to execute the task based on the plan and task type from the planner agent. 

We encapsulate the API-based LLM invocation behavior into a lightweight ROS package. Within the ROS system, the $llm$ node publishes instruction information from the API, which is subscribed to by the EgoPlanner algorithm to determine whether to initiate the path planning process, this framework enables monocular real-time obstacle avoidance (i.e., path planning functionality). In the LLM prompt for the execution agent, we instruct the LLM to employ pure agent functionality for independent tasks, converting naturally described tasks into machine language (code), and to invoke predefined tools through ROS for tool-assisted tasks Figure~\ref{fig:3}. To evaluate the performance of the execution agent in terms of Task Execution Success Rate (ESR) and UAV Energy Consumption (UEC), we will conduct experimental tests using the constructed UAV-GPT database.

For example, regarding a task ``Move forward 5 meters then take pictures of the kitchen and two bedrooms and avoid obstacles in time", assuming the system prompt indicates that there are obstacles at the doorways of the kitchen and the bedrooms, the execution process of the system is as follows:
\begin{enumerate}
    \item \textbf{Knowledge Matching \& Task Classification} 
    \begin{itemize}
        \item \textbf{Keyword Extraction:}
        \begin{itemize}
            \item Actions: move forward, take pictures, avoid obstacles
            \item Targets: kitchen, bedrooms
        \end{itemize}
        \item \textbf{Knowledge Base Comparison:}
        \begin{itemize}
            \item \textit{System Knowledge:} Predefined navigation parameters (e.g., egoplanner node), camera APIs.
            \item \textit{Internet Knowledge:} UAV Control APIs.
        \end{itemize}
        \item \textbf{Decision Mechanism:}
        \begin{itemize}
            \item Tool-Assisted Task : avoid obstacles requires real-time obstacle data → Triggers external collision-avoidance module.
        \end{itemize}
    \end{itemize}
    \item \textbf{Executable File Generation}
    \begin{itemize}
        \item execute\_egoplanner(target="kitchen", obstacle\_data=env\_obstacles)
        \item activate\_camera(camera\_type="RGB")  
    \end{itemize}
\end{enumerate}

\section{Experiment}
This section presents the experimental results of the UAV-GPT framework, aiming to validate its practicality through the operations of IRA, ESR, and UEC. Meanwhile, we have set up experimental scenarios in both simulation and real-world environments. The simulation scenario is built using the Airsim simulator and the Ue4 engine, while the real-world scenario is established with a laptop equipped with RTX3080 and a Tello UAV.

\subsection{Dataset and Evaluation Metrics}

\begin{figure}[H]    
    \centering  
    \includegraphics[width=0.8\linewidth]{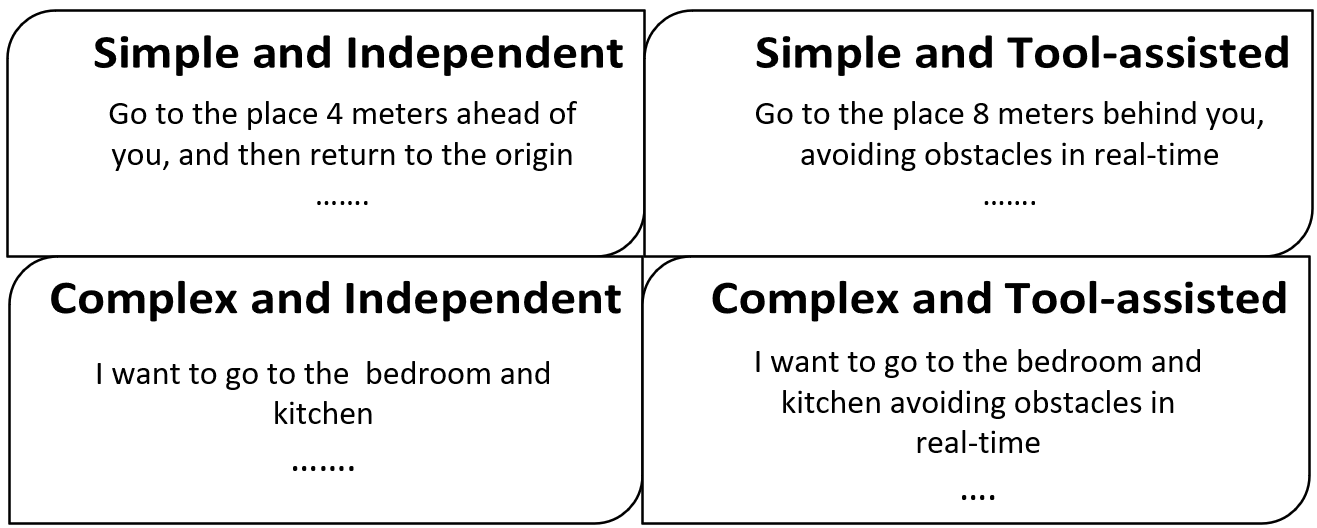}  
    \caption{\label{fig5}Datasets, the task here are only symbolic; in reality, the task classification is determined by the empirical parameters derived from the expert-annotated dataset.} 
\end{figure}

To evaluate the performance of our framework in complex tasks, we generated various requests that combine simple tasks Figure~\ref{fig5}, encompassing multiple decision types related to UAV control behaviors in Table~\ref{tab2}. Specifically, our dataset includes different experimental backgrounds such as moving forward, taking photos, searching for objects, indoor navigation, outdoor inspection, etc., as well as specific experimental tasks based on these backgrounds. Additionally, these tasks are expressed explicitly and implicitly. In explicit expressions, tasks clearly require the UAV to go somewhere and perform certain tasks, such as the task types proposed in Task List 1. However, implicit expressions do not explicitly state the purpose but use alternative phrasing to allow the LLM to understand 
the specific details of the task, for example, ``I want to eat watermelon, but I don't know where it is." The dataset contains a total of 160 task types for calculating $\mathcal{C}_{\text{task}}$, with each two-dimensional label containing 40 tasks. By comparing the task classification labels output by the planning end with the labels of the tasks themselves, IRA can serve as an effective measure of accuracy.

Furthermore, each task in the dataset has corresponding completion criterias, which are used to compare with the output of UAV-GPT's solutions to determine whether the agent has successfully addressed the users' needs. IRA measures the proportion of tasks accurately classified by the planning agent within a hypothetical task database, with human classification accuracy set as the 100\% benchmark. ESR reflects the ratio of successful task executions to total executions under the same task classification and plan inputs. Finally, UEC indirectly indicates power consumption by measuring the flight duration of the UAV during various task executions, we compare the UEC levels under the same task to indicate the energy-saving performance. Since the flight speed of the UAV is fixed in this scenario, we use flight time to estimate the UAV's energy consumption, enabling more precise data information.

\subsection{Prompt Engineering}
\begin{figure}[H]    
    \centering  
    \includegraphics[width=0.95\linewidth]{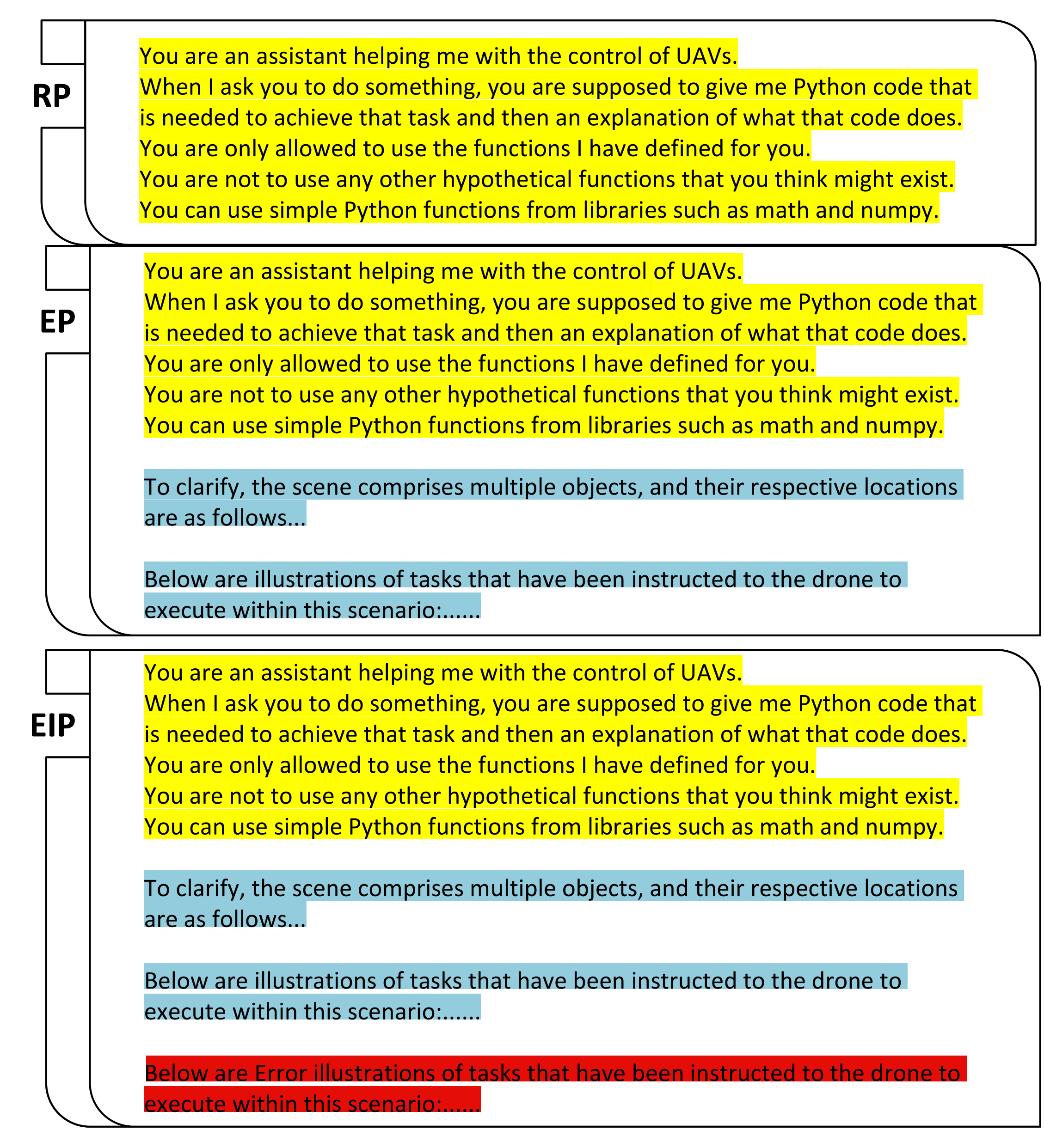}  
    \caption{\label{fig6}Prompts} 
\end{figure}

We briefly present different pre-prompt structures in Figure~\ref{fig6}

\textbf{General Prompt (RP)}:
In general prompts, the LLM is informed of the role it needs to play, the tasks it needs to perform, and the necessary location information for executing those tasks. However, it is not provided with any background knowledge related to the tasks or additional information about the current scenario. It needs to plan and act based on very limited cognition.

\textbf{Contextual Prompt (CP)}:
Based on general prompts, with contextual prompts, the LLM is provided with multiple task templates and corresponding outcomes for learning. Additionally, it is given extra information related to the input task, such as the coordinates of all objects in the scenario and the locations of potential obstacles. This information greatly enhances the LLM's ability to understand the scenario. We hope that through this approach, the LLM can accurately understand the user's needs and generate solutions.

\textbf{Iterative Prompt for Errors (EIP)}:
Based on the first two types of prompts, we simplify the failed solutions output by the LLM and use them as prompt texts in the pre-prompting stage. This can further refine the system prompts and improve its performance. By standardizing the structure and content of the prompts, this method can increase the number of task templates and outcomes (negative templates) in the prompts, allowing the LLM to anticipate more potential task scenario solutions and ensure more accurate and reliable classification and solution outputs.

\subsection{Experimental Results}
\label{4.3}
We used ERNIE-4.0 and GPT-4o as the base models and set the randomness of the models to 0 to eliminate possible random responses and ensure consistent classification of requests by the LLM. At the same time, we used GPT-3 and Llama3 70B as controls to enhance the persuasiveness of the experiments and verify the versatility of the framework. 
Firstly, Figure~\ref{fig7} clearly reveals the differences in task planning and execution strategies between our framework and the traditional HUI framework based on LLMs when it comes to executing tool-assisted tasks. Due to the incorporation of a ROS-based obstacle avoidance tool, UAV-GPT will plan a smoother route (b) when performing such tasks compared to the traditional framework. Moreover, because the dual-end agent can prevent LLMs from confusing planning with execution output, which may lead to irrational planning, UAV-GPT will devise a more reasonable and time-saving route when confronted with complex tasks (c). Therefore, we can qualitatively conclude that UAV-GPT exhibits better performance than the traditional LLM framework when dealing with tool-assisted and complex tasks. We repeated the experiments in Figure~\ref{figo} using the Tello UAV in real-world scenarios to further validate the reliability of our conclusions in Figure~\ref{figk}.

Next, in order to comprehensively evaluate the performance of UAV-GPT, we will adopt three indicators: IRA, ESR, and UEC, as the testing standards to quantitatively assess the performance of our framework:

\begin{figure}[H]  
    \centering  
    \includegraphics[width=1\linewidth]{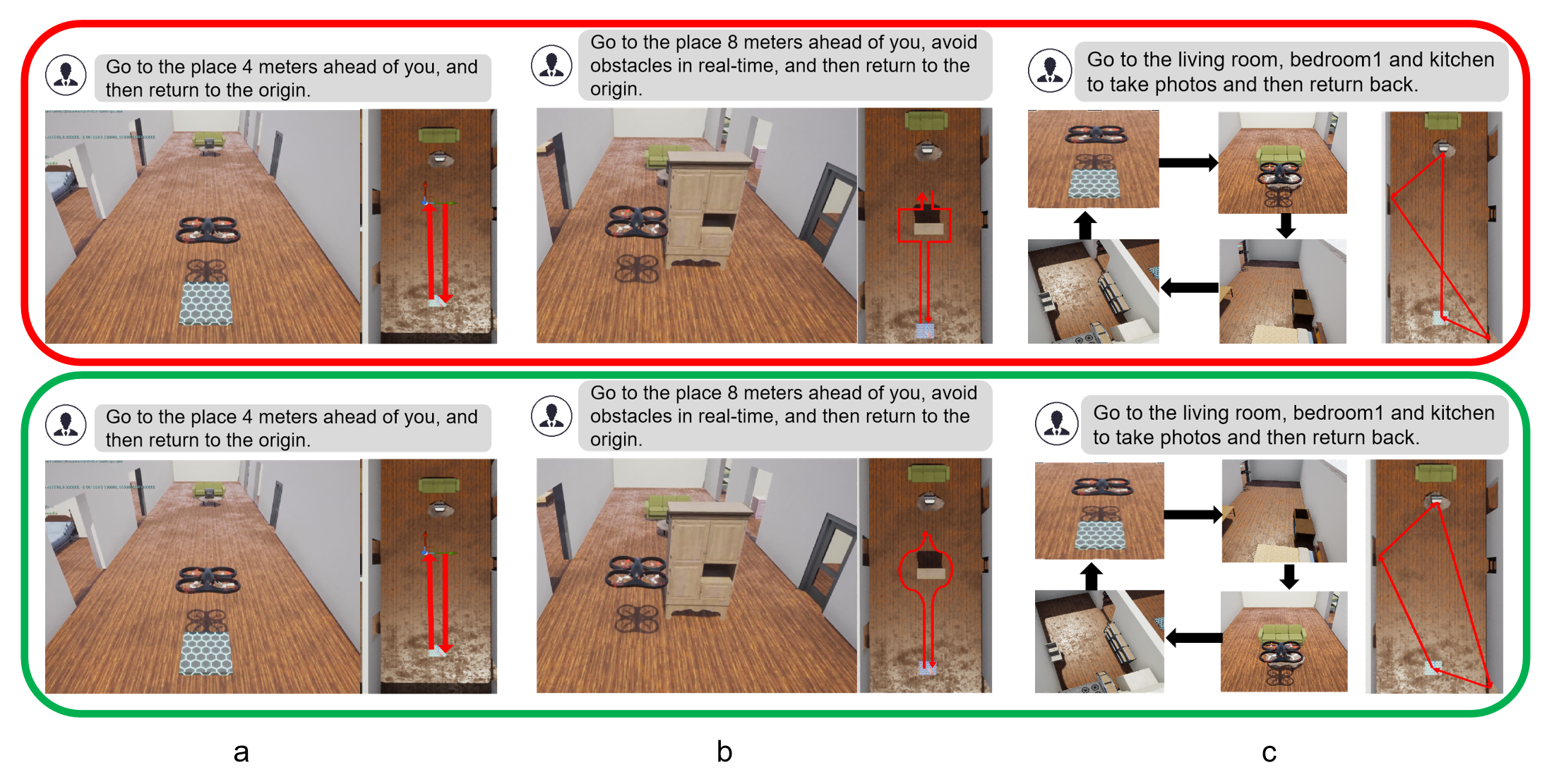}  
    \caption{\label{fig7}The single-ended planning framework (red box) and the UAV-GPT method (green box) exhibit significant differences in strategies when executing ST and CI tasks. UAV-GPT is more reasonable in execution.} 
\end{figure}

\begin{figure}[H]  
    \centering  
    \includegraphics[width=0.6\linewidth]{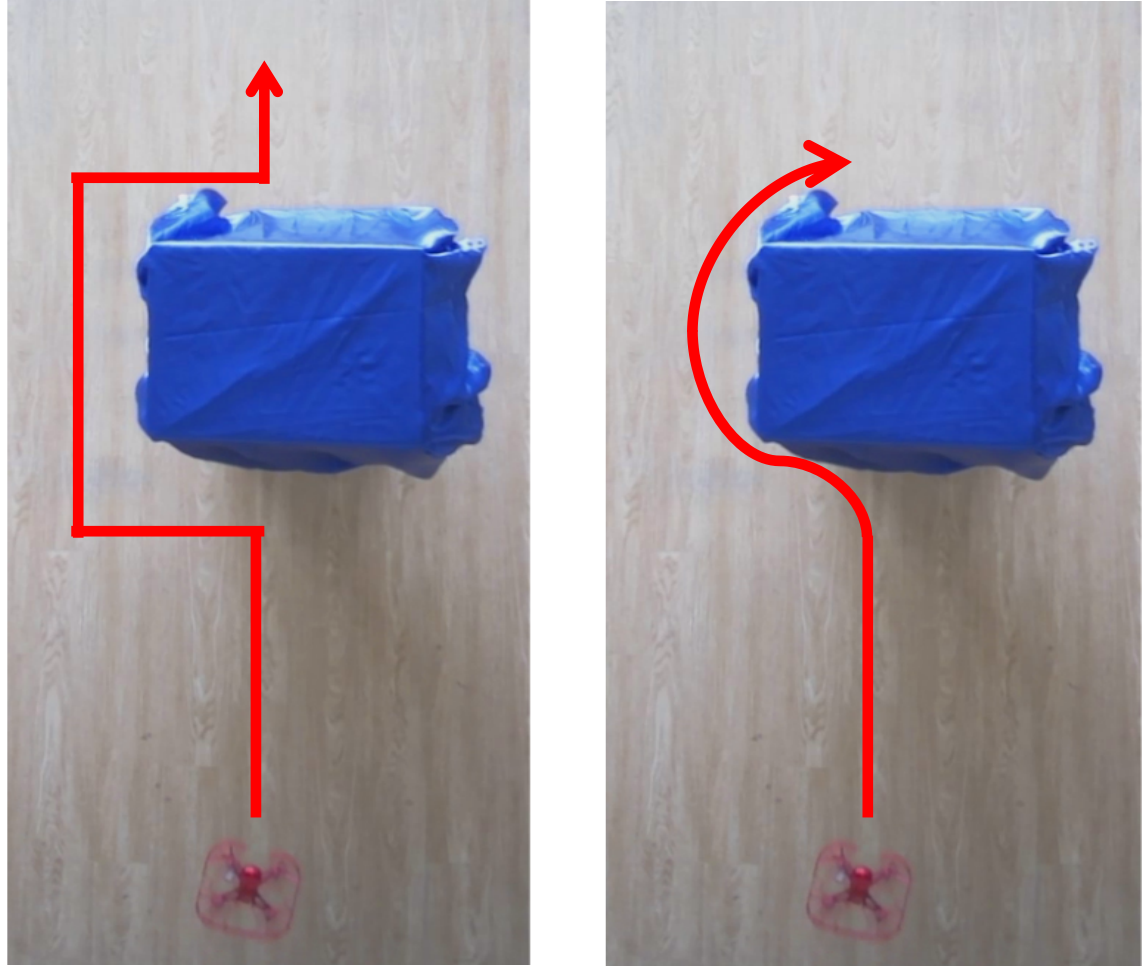}  
    \caption{\label{figo}This is the same tasks that we completed in real-world scenarios using the Tello drone} 
\end{figure}

\subsubsection{Intent Recognition Accuracy(IRA)}

\begin{figure}[H]  
      \centering  
      \includegraphics[width=0.5\linewidth]{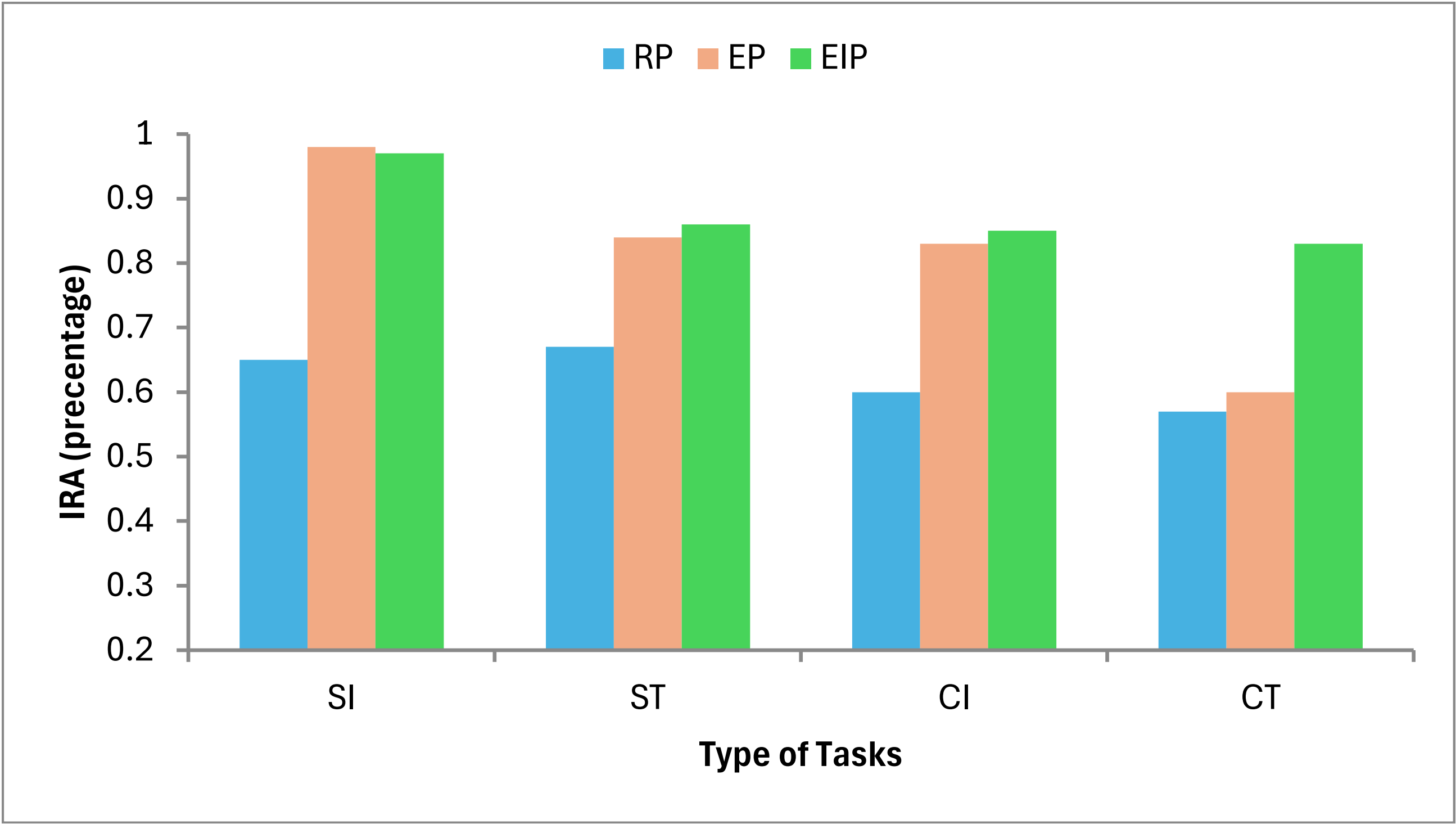}  
      \caption{\label{fig8}The performance comparison of IRA with different prompt.} 
\end{figure}

Figure~\ref{fig8} demonstrates the influence of employing various types of prefix prompts (EIP) on the success rate of task classification by UAV-GPT when confronted with tasks of diverse classification labels. Here, we uniformly use ERNIE-4.0 for testing to avoid the influence of different foundational LLMs on the success rate of prefix prompts in classification, as it provides a good balance between latency and accuracy and is therefore well-suited for our experiments. The findings reveal that EIP achieves the highest IRA metric among the three distinct types of prefix prompts. Specifically, in the CI and CT tasks, compared to RP, the success rate of EIP prompts has increased by an average of 25\%. When compared to EP, in the CT task, there is a 23\% improvement, which stems from the incorporation of environmental information and past misclassifications. Consequently, in subsequent experiments, we will adopt EIP prompts as the type of prefix prompt to minimize the influence of prefix prompts on the experimental outcomes. 

\subsubsection{Task Execution Success Rate(ESR)}
\begin{figure}[H]  
    \centering  
    \includegraphics[width=1\linewidth]{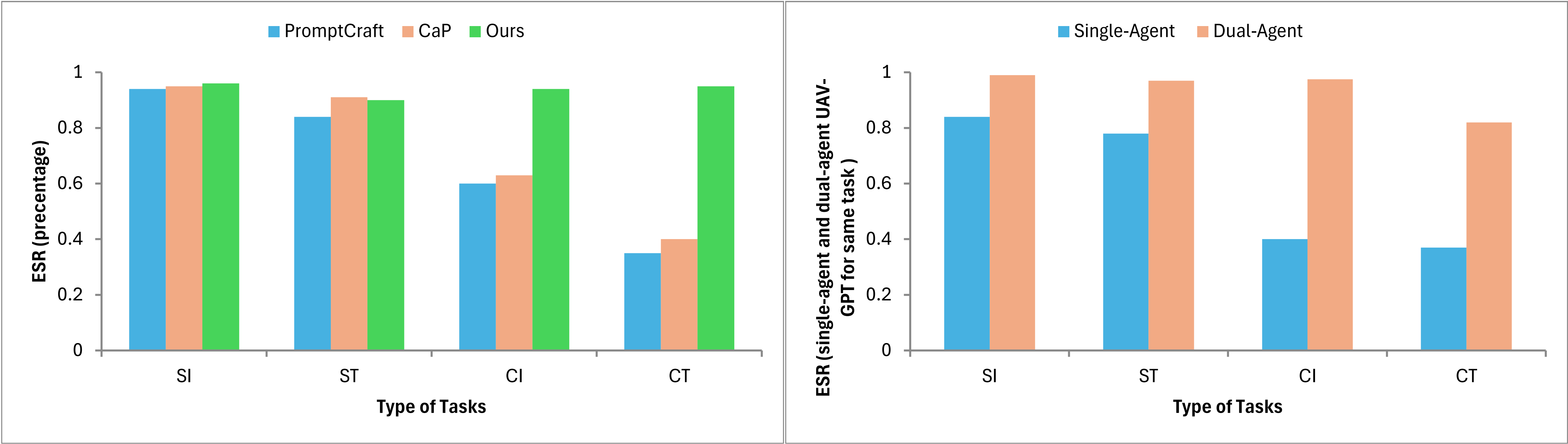}  
    \caption{The performance comparison of ESR with different HUI frameworks.}  
    \label{fig9}  
\end{figure}

Figure~\ref{fig9} demonstrates the differences in the response of three interaction frameworks to the ESR metric when faced with tasks of different classification labels, based on the premise of EIP. The results show that for simple independent tasks and simple tool tasks, there is no significant difference in success rate between the three methods; however, when faced with complex tasks, whether they are complex-simple tasks or complex-independent tasks, the introduction of the dual-end planner significantly increases the success rate of UAV-GPT compared to the other two methods (PromptCraft~\cite{vemprala2024chatgpt} and CaP~\cite{liang2023code}, representing single-agent LLM SOTA baselines). Additionally, `Ours' employs ERNIE-4.0 as the base LLM model. In the case of CI task classification, the efficiency of UAV-GPT improves by 34\% compared to the other two methods, and by 61\% in CT tasks. We attribute this result to the improvement in task classification efficiency and the addition of the planning end. To further explore the reasons, we conducted additional experiments on UAV-GPT, integrating the planning end and execution end into a single planner, and then performing task classification and execution. The results in Figure~\ref{fig9} confirm our hypothesis: for simple category tasks, there is no significant difference in success rate between single-agent planning and dual-agent planning; however, for complex category tasks, the advantage of dual-agent planning is evident.

\begin{figure}[H]  
      \centering  
      \includegraphics[width=0.5\linewidth]{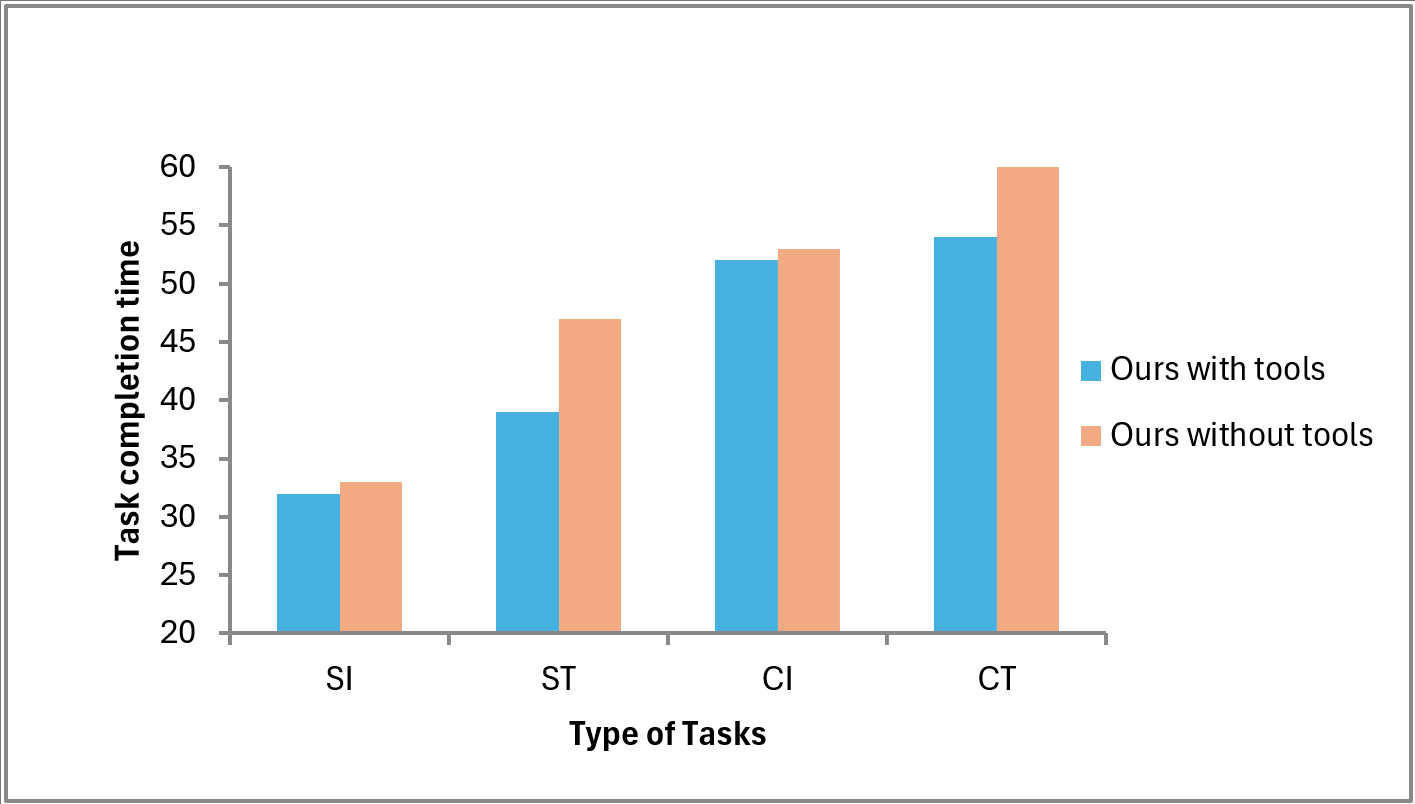}  
      \caption{\label{figk}The difference between an Execution Agent using external tools and not using external tools.} 
\end{figure}

We have also designed an experiment to verify whether the involvement of external tools can improve the success rate of task execution. We constructed a dataset consisting of 20 independent ``simple-tool" tasks and used UAV-GPT to execute these tasks under two different settings: one where the system can normally call external tools, keeping the task flow consistent with regular operation; the other where external tool calls are artificially prohibited, meaning that even if the system has identified the task as a ``simple-tool", it cannot actually schedule the relevant modules for execution. Through this comparative experiment, we aim to further verify the key role of external tools in task execution. The results in Figure~\ref{figk} clearly show that the system's execution capability is significantly affected in the absence of external tool support, thereby highlighting the importance of tool involvement for specific task types.

\subsubsection{UAV Energy Consumption (UEC)}
\begin{figure}[H]  
    \centering  
    \includegraphics[width=1\linewidth]{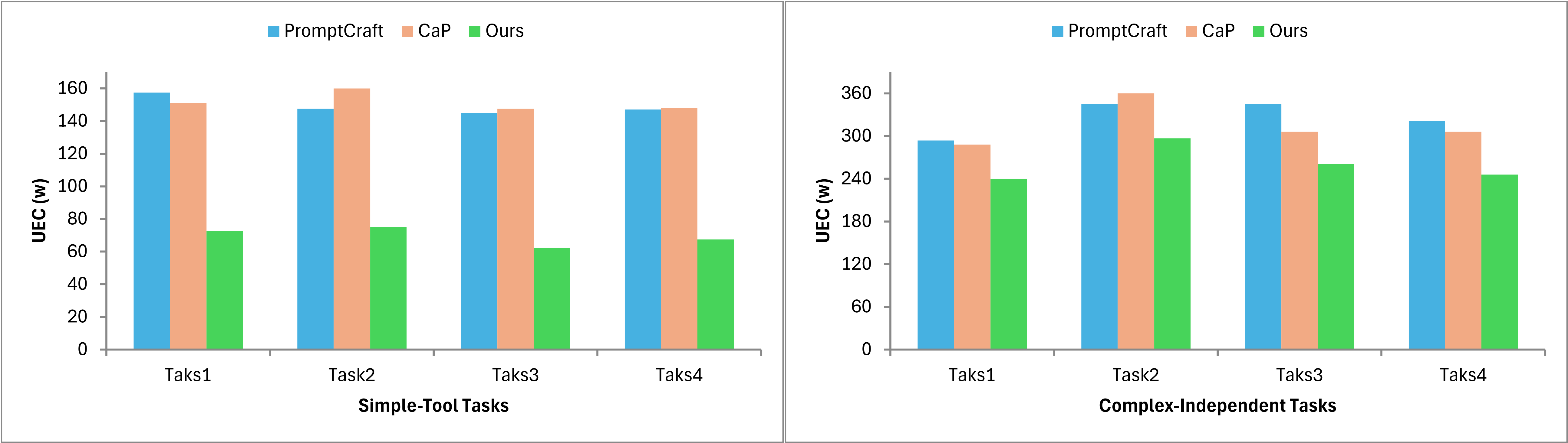}  
    \caption{The performance comparison of UEC with different tasks and HUI frameworks .}  
    \label{fig11}  
\end{figure}

Although UAV-GPT demonstrates a high completion rate across tasks with four classification labels, the efficiency of UAV task execution is also a crucial indicator that cannot be overlooked in practical applications. If a UAV can accomplish a task but takes an excessively long time, it is clearly unreasonable. Therefore, we next selected ST and CI category tasks for testing (we picked four ST and CI tasks from the first two indicator tests that all three interaction frameworks could complete), and let the three frameworks output solutions and execute them respectively. Figures~\ref{fig11} show that in the face of simple tool-based and complex independent category tasks, our method exhibits a significant advantage over the other two pure scripting and single-planning-end methods. To control variables and avoid confounding factors, we still used the EIP prompt—the best performer under the IRA metric—and fixed ERNIE-4.0 as the LLM base for all tested method. The PX4 Autopilot was also consistently used as the simulator UAV controller for three tested methods to minimize system-level interference.

\begin{figure}[H]  
    \centering  
    \includegraphics[width=0.8\linewidth]{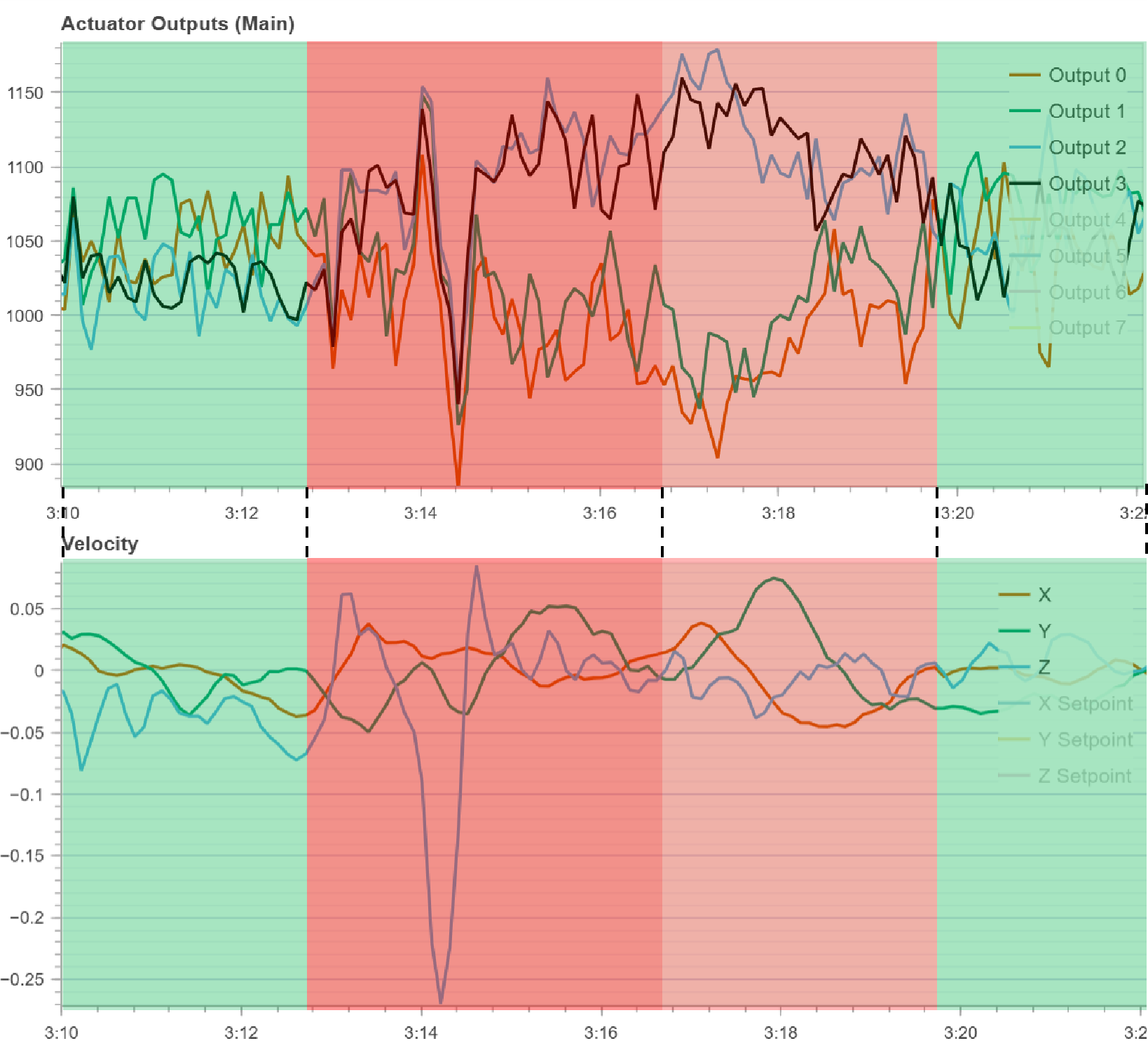}  
    \caption{In the chart, red areas represent the category of low-efficiency flights (low SPR values), and green areas represent the category of high-efficiency flights (high SPR values); different color depths are used to reflect the efficiency differences within the same category — the darker the red, the lower the flight efficiency.}  
    \label{pmw1}  
\end{figure}

In real-world scenarios, the power consumption of a drone before and after flight can be calculated from the battery capacity before and after the flight. However, there is no physical battery model in the simulation environment. Therefore, we analyze the ``Actuator Outputs (Main)” data in the simulated flight control to estimate the power consumption of the drone during task execution. As shown in the top picture of Figure~\ref{pmw1}, each curve corresponds to the PWM control signal of a main motor. Since the simulation does not model the physical loss of the motor, the PWM value is considered to represent the motor speed. According to the empirical physical formula, under the condition of constant air resistance and load, the propeller power $P$ and the motor speed $n$ have a cubic relationship:

\begin{equation}
    P \propto n^3
\end{equation}
thus, we can obtain the total power:
\begin{equation}
    P_{\text{total}}(t) = \sum_{i=1}^{4} P_i(t)
\end{equation}
finally, we will get the full power during one task:
\begin{equation}
    E_{\text{total}} \approx \int P_{\text{total}}(t) dt
\end{equation}

To more intuitively evaluate the energy consumption of the UAV during task execution, we introduced the ``Speed-to-Power Ratio (SPR)" as a visualization indicator on the basis of quantitatively calculating the UEC. This indicator is derived from the average vector velocity and quantized power consumption within each 2-second interval in the task scenario, and can clearly reflect whether the UAV's power consumption is used to increase speed rather than generating additional losses. We can calculate the SPR through this equation:
\begin{equation}
    \mathrm{SPR}_k = \frac{\|\vec{v}_k\|}{\sum_{t = t_k}^{t_k + 2s} P_{\text{total}}(t) \cdot \delta t}
\end{equation}
Where $k$ represents the $k$-th 2-second time interval, $\vec{v}_k$ is the average velocity vector within this interval, and $P_{\text{total}}(t)$ is the PWM-estimated power within this period of time.

Compared to PromptCraft and CaP, our method achieved an average performance advantage of 70w per task in ST tasks and 55w per task in CI tasks. This advantage stems from the integration of tool invocation and planning ends, enabling \sout{Large Language Models} LLMs to execute obstacle avoidance tasks in a more direct manner and plan paths from a global perspective without considering the details of specific execution codes.

To further analyze the efficiency changes during flight, we simultaneously monitored the effective operating range of the motor PWM and observed its fluctuation characteristics in the time series. In the Figure~\ref{pmw1}, the green area represents stable PWM output and balanced speed, indicating that the aircraft is in a stable propulsion state with high system operating efficiency; the red area shows sharp changes and high-frequency disturbances in the PWM signal, accompanied by speed jumps, which means the system is frequently adjusting the motor to maintain attitude or counter external disturbances, thereby leading to reduced efficiency and increased energy consumption. Based on the SPR value calculated from the vector velocity and quantized power consumption, we visualized the flight path in green (high efficiency) and red (low efficiency), We designed an interior space with three bedrooms, two living rooms, one kitchen and two bathrooms in Figure~\ref{room}. Next, we used two ST tasks to demonstrate the visualization of flight efficiency represented by SPR values, as shown in the Figure~\ref{room1} and Figure~\ref{room2}:

\begin{figure}[H]  
    \centering  
    \includegraphics[width=0.6\linewidth]{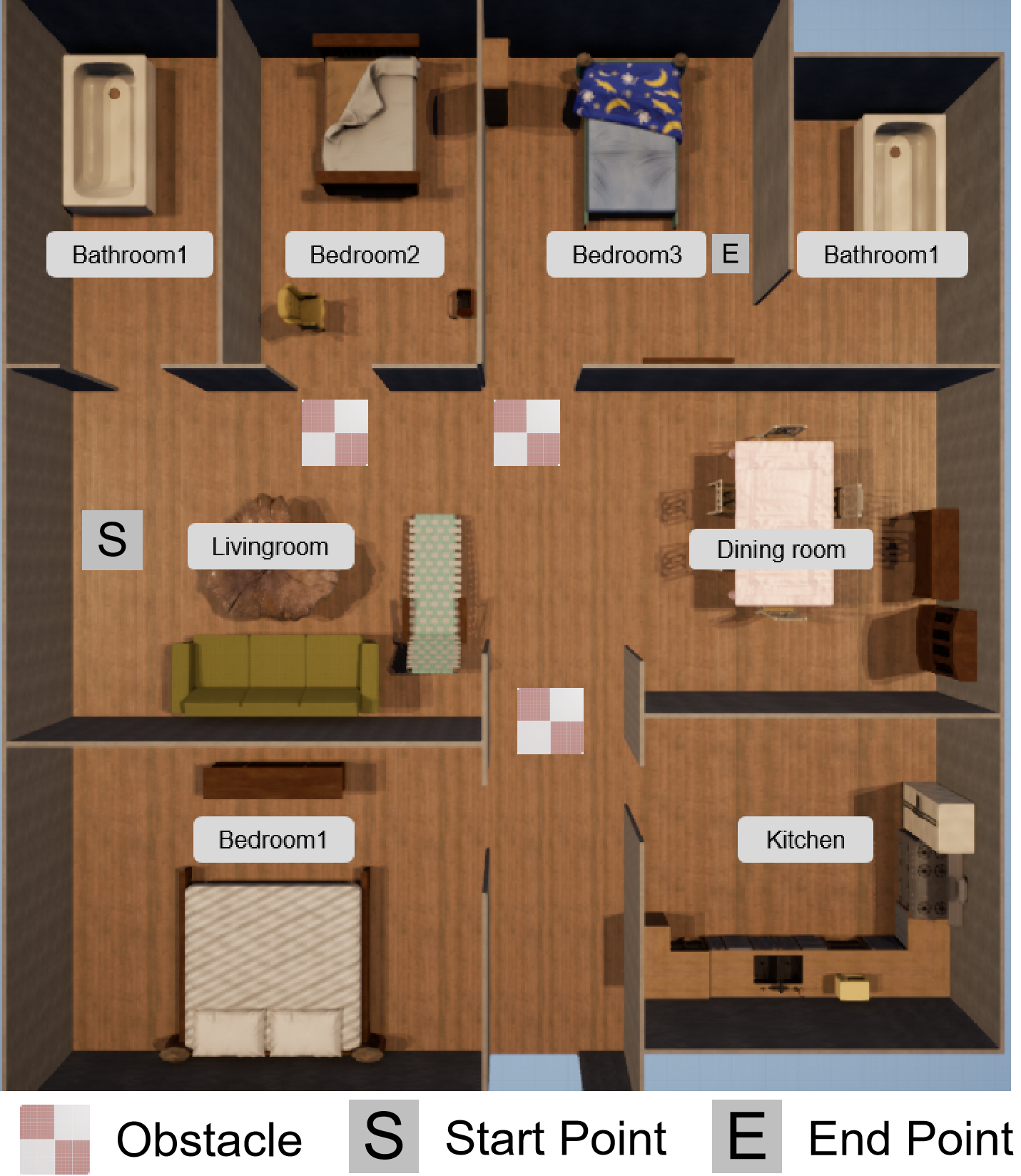}  
    \caption{Simulated indoor environment}
    \label{room}  
\end{figure}

\begin{figure}[H]  
    \centering  
    \includegraphics[width=0.8\linewidth]{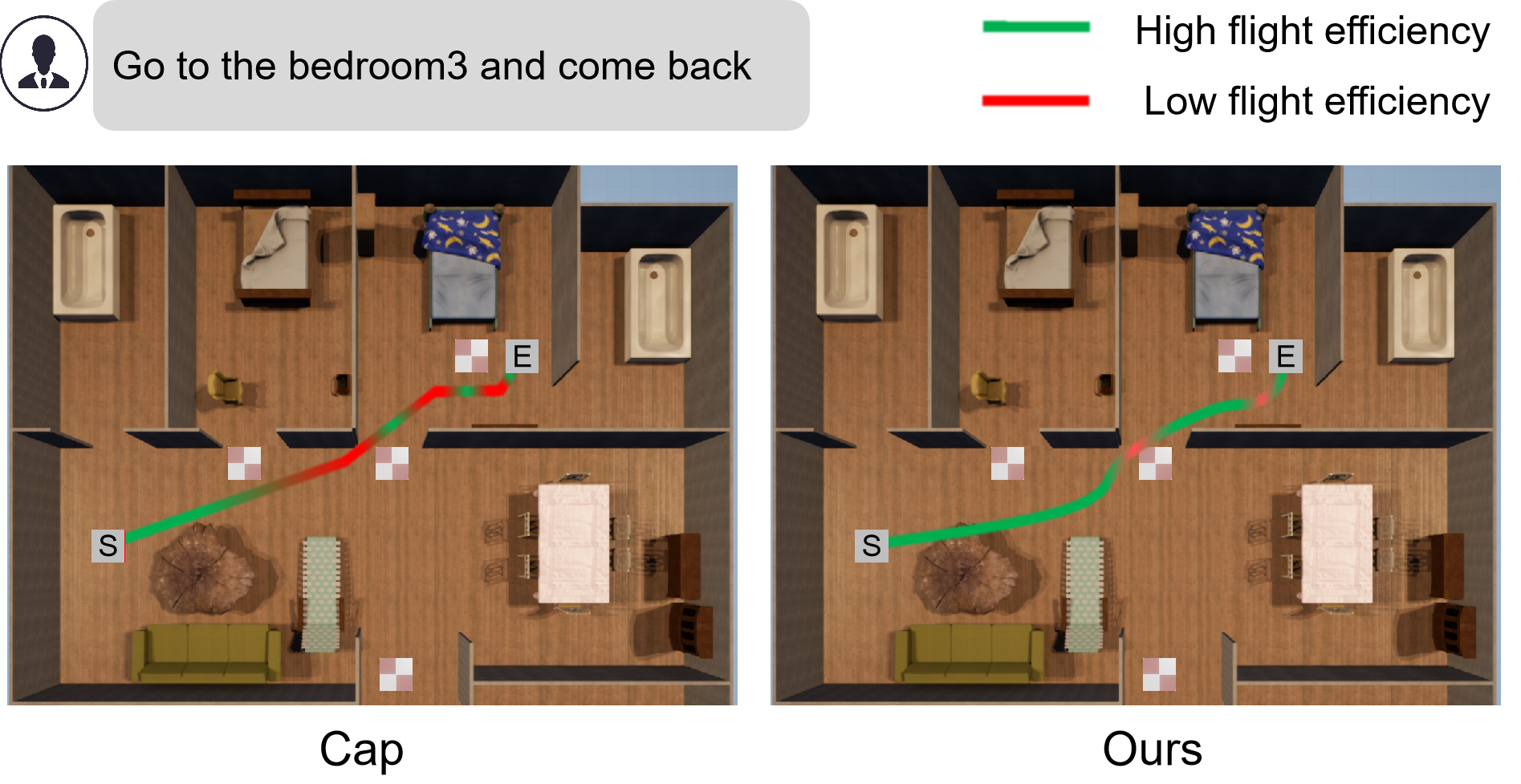}  
    \caption{Simple tool task 1}  
    \label{room1}  
\end{figure}

\begin{figure}[H]  
    \centering  
    \includegraphics[width=0.8\linewidth]{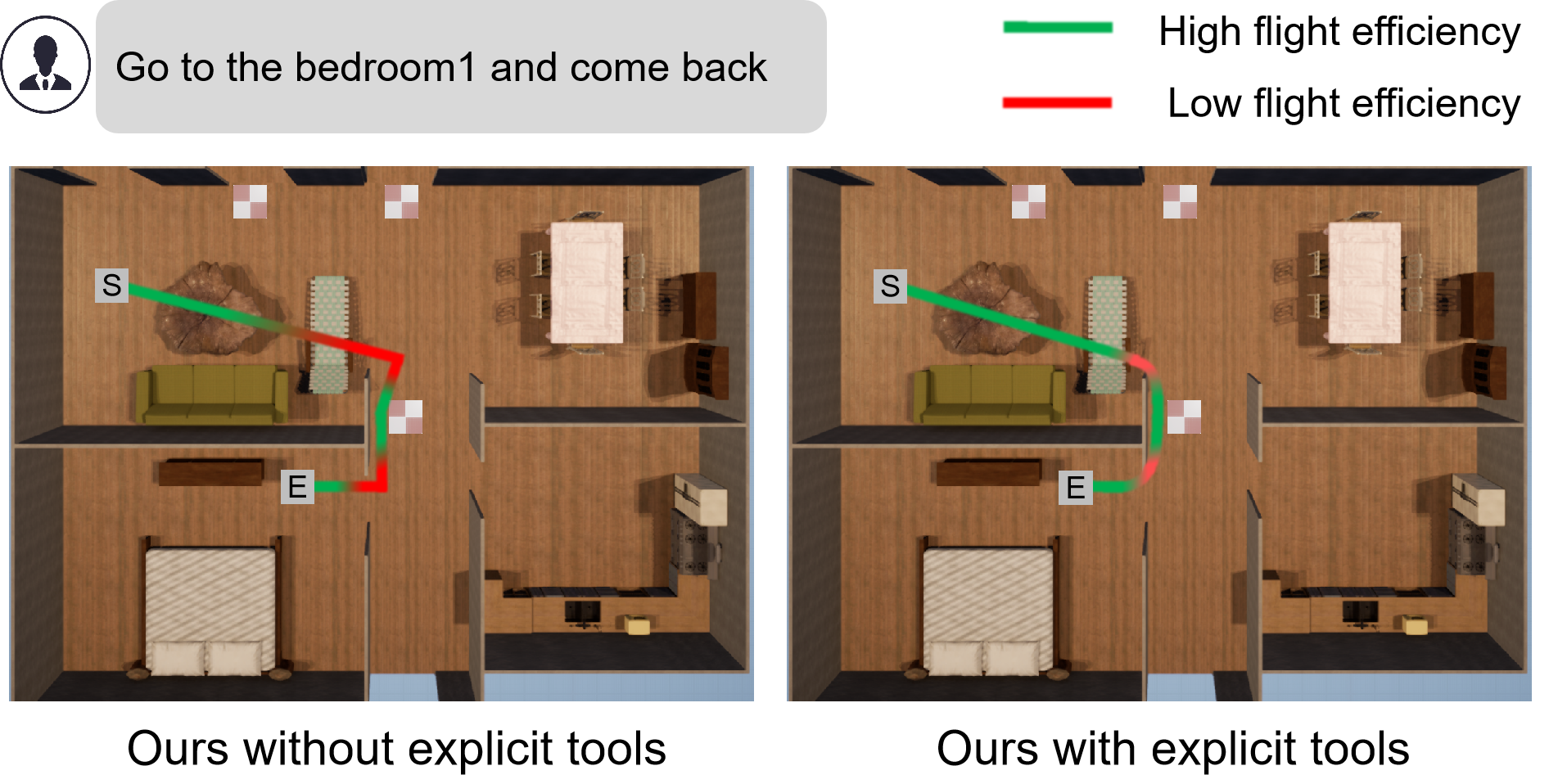} 
    \caption{Simple tool task 2}  
    \label{room2}  
\end{figure}

It is not difficult to find from the quantitative and visualized data that our agent system has achieved a significant improvement in the execution efficiency of ST and CI tasks. First, in terms of the planning and execution of CI tasks, the target sequence of high-level task planning is generated by large models. Since this paper compares a variety of different large model agents, and in order to ensure the fairness of the comparison and in line with the research direction of this paper, we have not carried out special tuning for a single large model in the comparison. It can be seen from the quantitative data that compared with the traditional single-end agent framework, the dual-end agent structure we adopted (which separates the top-level planning and bottom-level execution tasks) shows obvious advantages in the planning of task objectives. Second, it can be clearly seen from the Figure~\ref{room1} and Figure~\ref{room2} that in the ST task, compared with the case where only the large model at the single agent end is used for execution, the waypoints released by our framework during execution have better power efficiency. This smoother waypoint release effect is precisely due to the support of the dedicated tools we used.

We also evaluated the impact of different LLMs on the quality of work produced by this framework Table~\ref{tab3}. The models we selected include ERNIE-4.0, GPT-4o, GPT-3, and Liama3 70B. Using the prompt mode of EIP, we extracted 10 tasks from each of the four task categories in the dataset, totaling 40 tasks for classification testing. Table~\ref{tab3} shows the impact of different LLMs models on classification success rates: ERNIE-4.0 and GPT-4o performed the best, followed by Liama3 70B, and GPT-3 performed the worst.

\begin{table}[h]    
    \centering
    \caption{\label{tab3} The impact of different LLMs on the quality of work}
    \begin{tabular}{@{}>{\centering\arraybackslash}p{2cm}|
        >{\centering\arraybackslash}p{2cm}|
        >{\centering\arraybackslash}p{2cm}|
        >{\centering\arraybackslash}p{2cm}|
        >{\centering\arraybackslash}p{2cm}@{}}
    \hline
    & ERNIE-4.0 & GPT-4o&  Llama3 70B & GPT-3  \\ \hline
    Accuracy rate. &100\% &97\% &96\% &92\% \\\hline
    \end{tabular}
\end{table}

\subsection{User Study}
To evaluate the usability and user experience of the UAV-GPT framework in real HUI, we designed and conducted a user survey, inviting a total of 60 participants with professional backgrounds in drone or robot applications to take part in the experiment. We firstly aim to evaluate the user experience of the LLM-based UAV-GPT against the traditional ground station (GS) and secondly aim to compare our dual-agent framework with existing single-agent LLM frameworks (CaP and PromptCraft). To achieve this, we evaluated these two aspects separately which allow us to analyze the specific improvements of our dual-agent architecture over single-agent methods, without the results being confounded by the fundamentally different manual control paradigm.

We firstly constructed a user-study dataset by selecting five representative tasks from each of the four task classes in Section~\ref{3.1} (SI, ST, CI, CT), covering scenarios such as target inspection, area patrol and multi-step task planning, resulting in 20 tasks in total. For each chosen task, we generated four execution clips, each showing the task completed by: (1) GS, (2) UAV-GPT, (3) CaP and (4) PromptCraft. Participants were divided into two groups: Group 1 (n=20) was assigned to evaluate \textbf{Q1}, where they selected their preferred methods between UAV-GPT and GS, Group 2 (n=40) was assigned to evaluate the three LLM-based frameworks with respect to \textbf{Q2}: overall preference, \textbf{Q3}: which is easier to use, \textbf{Q4}: which is more adaptable to diverse environments, and \textbf{Q5}: which they would prefer in their professional domain. In both groups, participants viewed anonymized execution clips for 5 randomly chosen tasks from user-study dataset. Nest, we analyze the response data of user study.

As shown in Table~\ref{user1}, participants in Group 1 preferred the UAV-GPT over the GS across three task categories, demonstrating advantage in overall user experience. However, the system's reliance on LLM APIs introduces communication and inference latencies compared to direct manual control. This delay is particularly pronounced in complex tasks involving long action sequences, which partially attenuated the user preference for UAV-GPT in such scenarios. Furthermore, as indicated in Table~\ref{user2}, our method consistently outperformed CaP and PromptCraft across all four task categories in terms of \textbf{Q2}, \textbf{Q3}, \textbf{Q4} and \textbf{Q5}. The advantage narrowed slightly only in the lower-difficulty SI scenarios, where the selection proportions among the three methods were relatively closer.

\begin{figure}[H]  
    \centering  
    \includegraphics[width=0.8\linewidth]{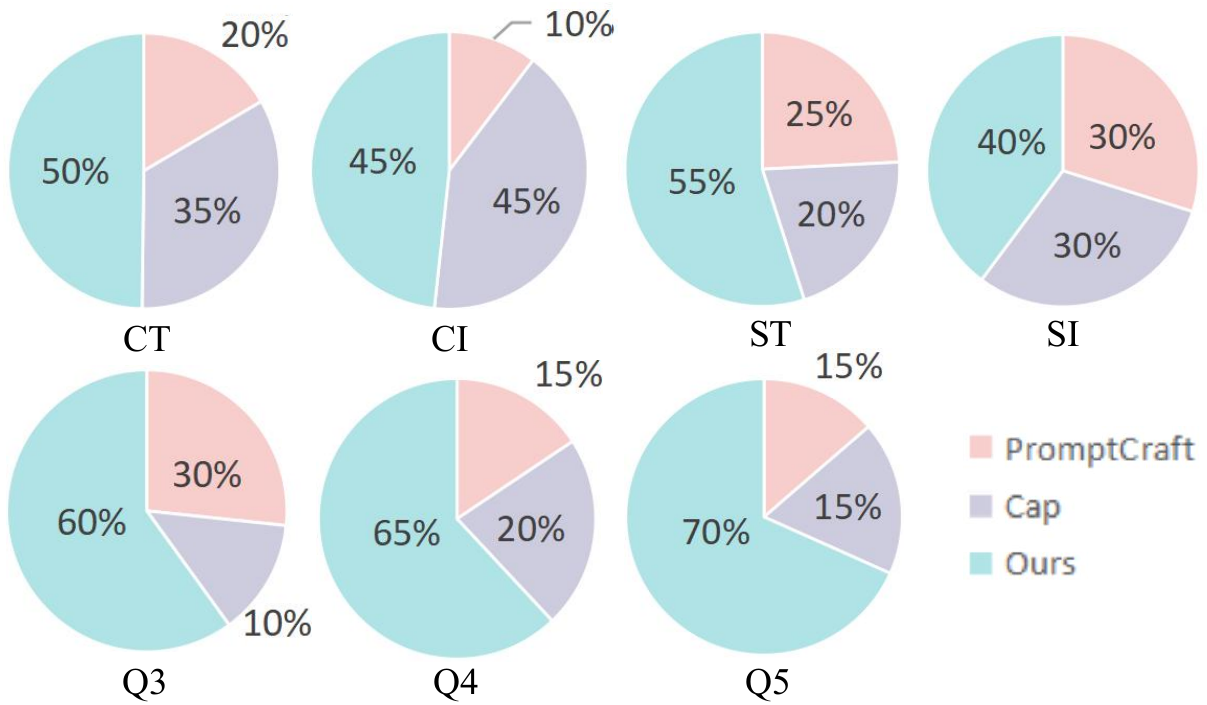}  
    \caption{Userstudy, this study clearly demonstrates that our method is more favored by users compared to the single-agent approach without external tools. }  
    \label{usr}  
\end{figure}

To statistically validate the user preferences observed in \textbf{Q2} (overall preference among LLM-based frameworks), we analyzed the original data from the 40 participants in Group 2. Each participant completed 5 trials, yielding 200 total observations. For the analysis, we quantified user preference by computing the selection proportion of each method (UAV-GPT, CaP, PromptCraft) for every participant in \textbf{Q2}. These proportions served as the dependent variable for a one-way repeated-measures ANOVA with \textit{method} as the within-subject factor.The analysis revealed a significant main effect of the framework method on user preference ($F(2, 78) = 13.12, p < 0.001$). Post-hoc pairwise comparisons indicated that UAV-GPT was chosen significantly more frequently than PromptCraft ($p < 0.001$) and showed a good preference advantage over CaP ($p \approx 0.05$). Additionally, CaP was selected more often than PromptCraft ($p < 0.05$). These statistical findings corroborate the descriptive distributions shown in Table~\ref{user2}, confirming that participants overall prefered UAV-GPT framework over other two methods. To provide a more intuitive presentation of the experimental results, we converted Table~\ref{user2} into a pie chart as shown in Figure~\ref{usr}.

\begin{table}[t]
\centering
\caption{\label{user1}Preference between UAV-GPT and GS in Group~1 (Q1).}
\begin{tabular}{c|c|c|c|c}
\hline
        & CT & CI & ST & SI \\\hline
GS      & 13 & 12 &  9 &  11 \\
UAV-GPT & 11 & 14 & 16 & 14 \\\hline
\end{tabular}
\end{table}

\begin{table}[t]
\centering
\caption{\label{user2}{red}{User preferences for LLM-based frameworks in Group~2 (counts of choices).}}
\begin{tabular}{c|c|c|c|c|c|c|c}
\hline
Method      & CT & CI & ST & SI & Q3 & Q4 & Q5 \\\hline
UAV-GPT     & 24 & 23 & 27 & 21 & 120 & 130 & 140 \\
CaP         & 16 & 23 & 10 & 16 &  20 &  40 &  30 \\
PromptCraft &  7 &  5 & 12 & 16 &  60 &  30 &  30 \\\hline
\end{tabular}
\end{table}

\section{Discussion and Limitations}

While UAV-GPT demonstrates significant improvements in task success rates (45.5\% ESR gain) and energy efficiency (62.5w UEC reduction per task) compared to single-agent frameworks, several limitations warrant discussion. The parameter tuning process for task complexity coefficients ($\alpha$, $\beta$, $\gamma_{p}$, $\gamma_{d}$, $\gamma_{a}$) relies on empirical fitting from expert-annotated datasets (CLAD and BRMData). Although effective in controlled scenarios Section~\ref{4.3}, this approach may require manual recalibration when deployed in novel dynamic environments (e.g., agricultural fields with shifting wind patterns), potentially increasing deployment overhead.
Regarding tasks adaptability, our framework achieves 96\% success in obstacle avoidance (Figure~\ref{fig9}) by integrating real-time replanning tools like EgoPlanner. However, current validation is confined to indoor/static settings (Figure~\ref{fig7}). Extreme outdoor conditions—such as simultaneous multi-obstacle avoidance or strong electromagnetic interference—remain unaddressed. The dependency on ROS-based tools (Section~\ref{3.3}) further limits applicability to resource-constrained edge devices, a critical gap for field applications like logistics or environmental monitoring.

Future work will prioritize two directions: First, implementing federated learning for autonomous parameter tuning to reduce manual intervention. Second, extending validation to heterogeneous outdoor scenarios with lightweight toolchains.

\section{Conclusion}
This paper proposed the ``UAV-GPT" framework based on the UAV, aiming to achieve more natural and flexible HUI through LLMs. Traditional UAV interaction designs are limited by predefined task planning, making it difficult to meet users' personalized needs. UAV-GPT achieves direct conversion and execution of user voice or text requests by constructing two LLM agents: a task planning agent and an execution agent. This framework leveraged the natural language understanding capabilities of LLMs to accurately classify and plan complex tasks, intelligently selecting execution codes or invoking tools to tackle them. Experimental results show that UAV-GPT significantly enhanced the fluency of HUI and the flexibility of task execution. Compared to traditional single-ended LLM planning HUI methods, UAV-GPT achieved an average improvement of 24\% in Intent Recognition Accuracy, 45.5\% in Task Execution Success Rate, and a reduction of 62.5w in energy consumption per task, effectively meeting users' personalized demands.





\bibliography{sn-bibliography}

\end{document}